\documentclass[10pt,twocolumn,letterpaper]{article}

 \usepackage[pagenumbers]{cvpr} 
\usepackage{mathtools} 
\usepackage{amsmath,amsfonts,amsthm,bm,mathtools, bbm}
\usepackage{amsthm}
\usepackage{amssymb}
\usepackage{eqparbox}
\usepackage{arydshln}
\usepackage[vlined]{algorithm2e}
\usepackage{algorithmic}
\usepackage{multirow}
\usepackage[pagebackref,breaklinks,colorlinks,citecolor=cvprblue]{hyperref}
\newtheorem{theorem}{Theorem}[section]

\newtheorem{lemma}[theorem]{Lemma}

\usepackage{amsmath,amsfonts,bm}

\def\eqref#1{equation~\ref{#1}}

\def\1{\bm{1}}

\DeclareMathAlphabet{\mathsfit}{\encodingdefault}{\sfdefault}{m}{sl}
\SetMathAlphabet{\mathsfit}{bold}{\encodingdefault}{\sfdefault}{bx}{n}

\DeclareMathOperator*{\argmax}{arg\,max}
\DeclareMathOperator*{\argmin}{arg\,min}

\usepackage[dvipsnames]{xcolor}

\definecolor{cvprblue}{rgb}{0.21,0.49,0.74}
\usepackage[pagebackref,breaklinks,colorlinks,citecolor=cvprblue]{hyperref}

\usepackage{graphicx}
\usepackage{wrapfig}
\usepackage{amsmath} 
\usepackage{tabularx}
\usepackage{float}
\usepackage[export]{adjustbox}
\usepackage{multirow}
\usepackage{dsfont}
\usepackage{xcolor,colortbl}
\newtheorem{prop}{Proposition}
\usepackage{amssymb}
\usepackage{pifont}
\newcommand{\xmark}{\ding{55}}%

\usepackage{color} 

\newcommand*{\affaddr}[1]{#1} 
\newcommand*{\affmark}[1][*]{\textsuperscript{#1}}

\def\paperID{12768} 
\def\confName{CVPR}
\def\confYear{2024}

\title{LP++: A Surprisingly Strong Linear Probe for Few-Shot CLIP}

\author{%
    \begin{tabular}[t]{c}
        Yunshi Huang\affmark[1]\thanks{Equal contribution. Correspondence to \href{mailto:yunshi.huang@etsmtl.ca}{yunshi.huang@etsmtl.ca}, \href{mailto:fereshteh.shakeri.1@etsmtl.net}{fereshteh.shakeri.1@etsmtl.net}} \and 
        Fereshteh Shakeri\affmark[1]$^*$
    \end{tabular} \\
    \begin{tabular}[t]{c}
        Jose Dolz\affmark[1] \and
        Malik Boudiaf\affmark[1] \and
        Houda Bahig\affmark[2] \and
        Ismail Ben Ayed\affmark[1]
    \end{tabular} \\[1em]
    \affaddr{\affmark[1]\'ETS Montr\'eal},
    \affaddr{\affmark[2]Université de Montréal}\\
}

\begin{document}
\maketitle
\begin{abstract}

In a recent, strongly emergent literature on few-shot CLIP adaptation, Linear Probe (LP) has been often reported as a weak baseline. This has motivated intensive research 
building convoluted prompt learning or feature adaptation strategies. In this work, we propose and examine from convex-optimization perspectives a generalization of the standard LP baseline, in which the linear classifier weights are learnable functions of the text embedding, with class-wise multipliers blending image and text knowledge. As our objective function depends on two types of variables, i.e., the class visual prototypes and the learnable blending parameters, we propose a computationally efficient block coordinate Majorize-Minimize (MM) descent algorithm. In our full-batch MM optimizer, which we coin LP++, step sizes are implicit, unlike standard gradient descent practices where learning rates are intensively searched over validation sets.
By examining the mathematical properties of our loss (e.g., Lipschitz gradient continuity), we build majorizing functions yielding data-driven learning rates and derive approximations of the loss's minima, which provide data-informed initialization of the variables. Our image-language objective function, along with these non-trivial optimization insights and ingredients, yields, surprisingly, highly competitive few-shot CLIP performances. Furthermore, LP++ operates in black-box, relaxes intensive validation searches for the optimization hyper-parameters, and runs orders-of-magnitudes faster than state-of-the-art few-shot CLIP adaptation methods. Our code is available at: \url{https://github.com/FereshteShakeri/FewShot-CLIP-Strong-Baseline.git}.

\end{abstract}    
\section{Introduction}
\label{sec:intro}

\begin{figure}
     \centering
     \includegraphics[width=0.9\linewidth]{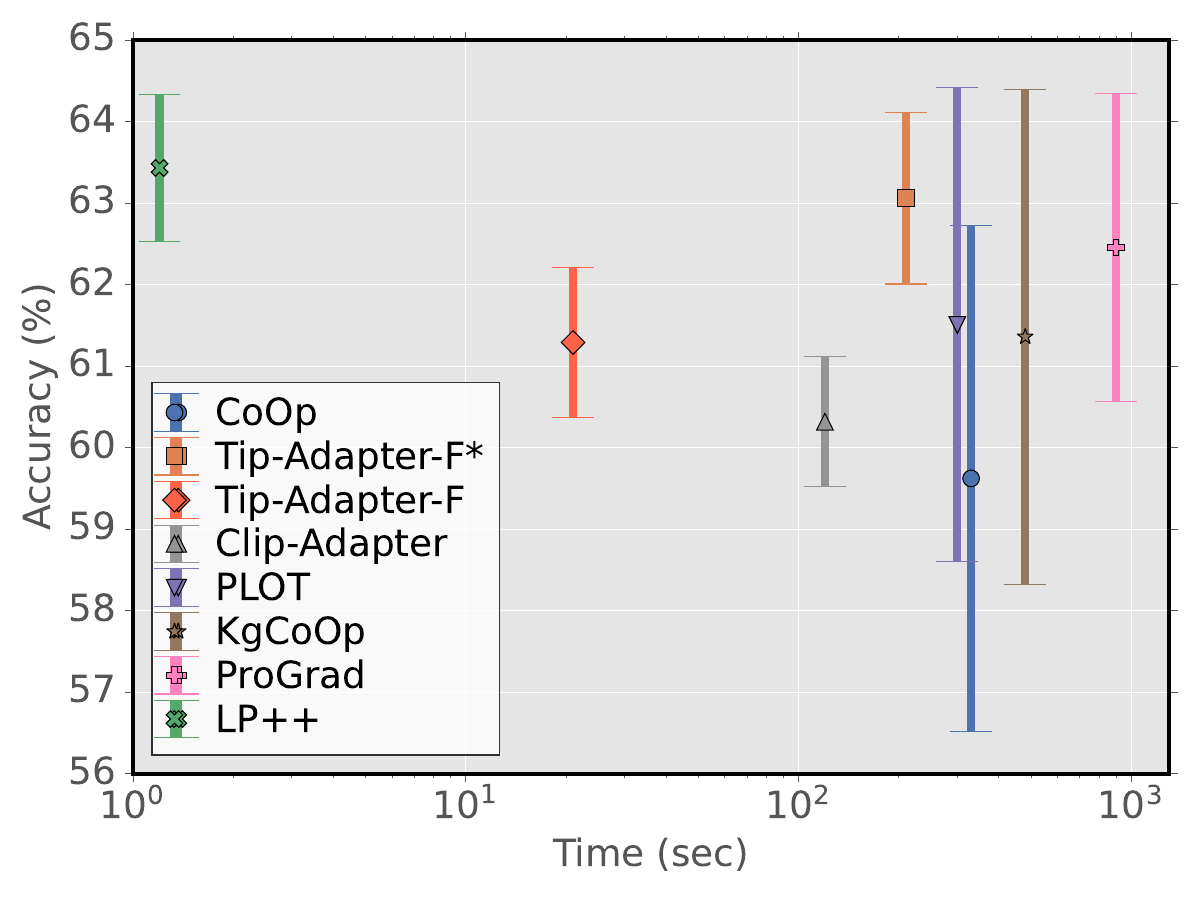}
    \caption{ Comparison of LP++ with state-of-the-art few-shot CLIP methods in the 1-shot setting across 11 datasets. We compute the mean accuracy and standard deviation using 10 random tasks for each dataset. The error bars indicate the average standard deviation over all 11 datasets. 
    The x-axis represents 
    the run time for one task, 
    averaged over the 11 datasets.
    Tip-Adapter-F and Tip-Adapter-F$^*$ are two re-implementations of Tip-Adapter-F \cite{zhang2022tip}, with fixed and grid-search hyper-parameters, respectively (implementation details provided in \cref{sec:baselines}).}
    \label{fig:avg-std}
\end{figure}

Recently, there has been a growing popularity in multimodal learning methods, which process and merge information from diverse modalities. In particular, large-scale vision-language models (VLMs), such as CLIP~\cite{radford2021clip} and ALIGN~\cite{jia2021scaling}, have attracted wide attention and made substantial progress in computer vision, showing promising generalization capabilities in various downstream tasks. Unlike conventional task-specific models that are trained with a predetermined set of labels, these so-called {\em foundation} models learn to align images with text, in an open-vocabulary fashion. They train, via contrastive learning, vision and text embeddings jointly using a large-scale amount of image-text pairs collected over the internet, thereby leveraging the rich semantic knowledge inherent to language (e.g., concept hierarchies). For a given downstream image classification task, vision-language embeddings enable {\em zero-shot} predictions, without re-training, using textual descriptions of the classes (a.k.a {\em prompts}). For instance, for a given class $k$, a textual description of the class, which we denote $\bm{z}_k$, could 
be ``a photo of a [class$_k$]'', where [class$_k$] is the class name. Thus, the zero-shot class prediction for a query image $\bm{x}$ is obtained from the cosine similarity between the $l_2$-normalized vision-encoded embeddings, $\bm{f} = {\bm\theta}_v (\bm{x})$, and the text-encoded ones, $\bm{t}_k = {\bm\theta}_t (\bm{z}_k)$: $\hat{k} = \argmax_{k} \bm{f}^t \bm{t}_k$, where $t$ denotes the transpose\footnote{For $l_2$-normalized feature embeddings, the dot product corresponds to the cosine similarity.}.

Motivated by the observation that the choice of input prompts $\bm{z}_k$ may affect the zero-shot predictions, and following on from the strong recent emergence 
of {\em prompt learning} research in the NLP community \cite{ShinEMNLP2020,JiangACL2020,Hong2021NAACL}, the popular work in ~\cite{Zhou2022coop} pioneered {\em context optimization} (CoOp) for vision-language models. CoOp models input text $\bm{z}_k$ as learnable continuous vectors, e.g., in the form $\bm{z}_k=(\bm{z}_k^1, \dots, \bm{z}_k^M, [\mbox{class}_k])$, where $(\bm{z}_k^l)_{1 \leq l \leq M}$ are {\em learnable} text tokens, $[\mbox{class}_k]$ is a fixed token corresponding to the word embedding vector of the name of the $k^{th}$ class, and $M$ is a hyper-parameter. These learnable vectors are fine-tuned as task-specific prompts using few-shot training examples and a standard supervised classification loss. More specifically, in this few-shot setting, we assume access to a set consisting of a few labeled samples 
for each target class, often referred to as the {\em support set}. Let $\bm{f}_i = {\bm\theta}_v (\bm{x}_i)$ denote the vision embedding of support image $i$, and $y_{ik}$ its one-hot encoded label, 
i.e., $y_{ik}=1$ if image $\bm{x}_i$ belongs to class $k$ and $0$ otherwise. Expressing the text embeddings as $\bm{t}_k = {\bm\theta}_t (\bm{z}_k^1, \dots, \bm{z}_k^M, [\mbox{class}_k])$,  
CoOp fine-tunes text tokens $(\bm{z}_k^l)_{1 \leq l \leq M}$ by minimizing the cross-entropy (CE) loss, with $N$ labeled support samples and $K$ classes\footnote{The number of labeled support samples per class, $S=\frac{N}{K}$, is small, typically in $\{1, 2, 4, ..., 16\}$.}: 
\begin{equation}
\label{CE-loss-coop}
  - \frac{1}{N}\sum_{i=1}^{N}\sum_{k=1}^{K} y_{ik}\ln{p_{ik}}
\end{equation}
where the softmax predictions $p_{ik}$ and the logits (class scores) $l_{ik}$ are given by: 
\[p_{ik} =   \frac{\exp \left ( l_{ik} \right) }{\sum_{j=1}^{K}\exp \left ( l_{ij} \right )}; \quad l_{ik} = \bm{f}_{i}^{t} \bm{t}_{k} \]

Although recent, the pioneering idea of CoOp has triggered a quite  abundant literature on prompt learning for few-shot vision-language models, with numerous, more convoluted extensions, e.g.
 \cite{chen2023plot, Yao2023kgcoop, Zhu2023prograd}, to list a few. For instance, PLOT~\cite{chen2023plot} followed up by learning multiple prompts, to describe the characteristics of each class, via minimizing an optimal-transport distance. KgCoOp~\cite{Yao2023kgcoop} improves CoOp's performance when dealing with unseen classes, via minimizing the discrepancy between the text embeddings generated by the learned prompts and hand-crafted ones. While CoOp directly updates the context vectors using the CE loss, ProGrad \cite{Zhu2023prograd} aligns the few-shot downstream knowledge with the large-scale general knowledge, thus mitigating the overfitting of the few-shot samples.

Prompt learning methods have brought significant improvements over zero-shot classification, but they come at the price of heavy computational and memory load, as they require gradient back-propagation through the entire text encoder. Furthermore, they assume knowledge of the text encoder. These aspects may impede their deployment in low-resource and black-box, privacy-preserving scenarios, which are of wide interest in practice. Indeed, in NLP, there is currently an emerging literature on fast few-shot adaptation of black-box models \cite{ColomboEMNLP2023}, strongly motivated by the fact that large-scale foundation models (e.g., the GPT family, Anthropic’s Claude or Google’s PaLM) are only available through APIs and their pre-trained weights are not shared. Finally, by evaluating prompt learning methods over larger numbers of sampled support sets in our experiments (~\cref{fig:avg-std}), we observed that they exhibit large variation in performances. This could be explained by the fact that, through the text encoder, they learn prompts that are `'too specialized'' for a given image support set. 

While prompt learning alters the textual inputs, another category of approaches, referred to as {\em adapters}, focused on transforming the pre-training features 
of the visual or language encoders, e.g., \cite{Gao2023clipadapter,zhang2022tip}. These adapters are {\em non-linear} transformations, for instance, in the form of 
multi-layer modules, added to the encoder's bottleneck. They learn additional transformations, yielding logits of the form:
\begin{equation}
\label{logits-adapaters}
l_{ik}  = {\bm\theta}_a (\bm{f}_i,\bm{t}_k)
\end{equation}
The adapter's learnable parameters, $\bm{\theta}_a$, are fine-tuned over a few-shot task by optimizing the cross-entropy loss, similarly to (\ref{CE-loss-coop}) but with logits $l_{ik}$ expressed as functions of ${\bm\theta}_a$. For instance, the popular CLIP-Adapter \cite{Gao2023clipadapter} integrated a multi-layered perceptron to modify the features, along with residual connections, which enable blending with the original pre-trained features. Tip-Adapter \cite{zhang2022tip} added a non-linear, quadratic-complexity module, which 
evaluates the pairwise similarities between the features of the support sets, and blends the ensuing class scores with the original textual features. This category of approaches 
mitigates some of the limitations of prompt-learning methods as they result in few-shot adaptation that has significantly lower computation and memory loads. However, as shown in Fig. \ref{fig:avg-std} and in our experiments, their performances seem to depend strongly on some key hyper-parameters that have to be adjusted carefully on each downstream task, e.g. those 
that control the blending between the vision and language features. Therefore, to perform competitively, they incur an additional computation overhead to the adaptation phase, due to 
intensive (e.g., grid) search of the hyper-parameters over task-dedicated validation sets.

In the above-mentioned, strongly emergent literature on few-shot CLIP adaptation, {\em linear probe} (LP) \cite{radford2021clip} has been often reported as a very weak baseline. 
For instance, in the 1-shot setting, it scores near $20\%$ lower than the zero-shot predictions averaged over 11 benchmarks (Table \ref{table:resultOther_AllDatasets}). 
Initially evaluated in \cite{radford2021clip}, LP is a linear classifier on the vision-encoded features. Specifically, it optimizes the CE loss (\ref{CE-loss-coop}) w.r.t the last-layer weights of the vision encoder (i.e., the class prototypes), which we will denote $(\bm{w}_{k})_{1 \leq k \leq K}$ in the rest of the paper, with the logits given by: $l_{ik} = \bm{f}_{i}^{t} \bm{w}_{k}$. {\em A clear deficiency in this standard LP baseline is that it omits completely the language knowledge of CLIP, i.e.,} $(\bm{t}_k)_{1 \leq k \leq K}$. 

In this work, we propose and examine from convex-optimization perspectives a generalization of the standard LP baseline. Specifically, we extend the logits in the CE loss in (\ref{CE-loss-coop}), so that they become learnable functions of the text embedding: 
\[l_{ik} = \bm{f}_{i}^{t} (\bm{w}_{k} + \alpha_k \bm{t}_{k}) \]
with $(\alpha_k)_{1 \leq k \leq K}$ trainable class-wise parameters blending image and text knowledge. 
As our objective function depends on two types of variables, i.e., the visual class prototypes
$(\bm{w}_{k})_{1 \leq k \leq K}$ and blending parameters $(\alpha_k)_{1 \leq k \leq K}$, we propose a computationally efficient Block Majorize-Minimize (BMM) procedure. In our full-batch MM optimizer, which we coin LP++, step sizes are implicit in the definition of the majorizing functions, unlike standard gradient descent practices where learning rates are intensively searched over validation sets. Moreover, we examine the mathematical properties of our objective, i.e., (i) Lipschitz gradient continuity and (ii) decomposition into convex functions having closed-form optima. This enabled us to build majorizing functions yielding data-driven learning rates, and to derive approximations of the objective-function minima, which yield data-informed initializations of the variables. Our image-language objective function, along with these non-trivial optimization insights and ingredients, yield, surprisingly, highly competitive few-shot CLIP performances (Fig.~\ref{fig:avg-std}). Furthermore, LP++ operates in black-box, relaxes intensive validation searches for the optimization of hyper-parameters, and runs orders-of-magnitudes faster than state-of-the-art few-shot CLIP methods (Table~\ref{table:time_Imagenet}). For instance, for 16-shot ImageNet adaptation, it takes seconds on a single NVIDIA RTX A600 GPU.

\section{Formulation of LP++}
\label{method}

\begin{figure}
     \centering
    \includegraphics[width=\linewidth]{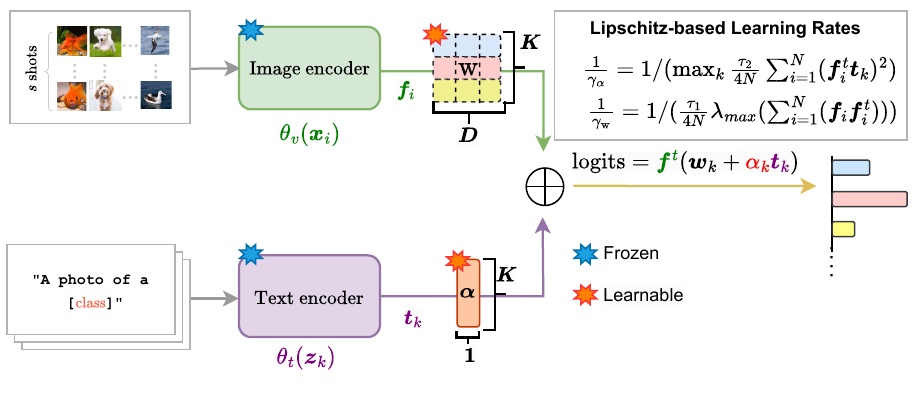}
    \caption{\textbf{Visualization of LP++.}}
    \label{fig:LP++}
\end{figure}

Following the notations introduced in the previous section, we propose to minimize the following CE objective function w.r.t
visual class prototypes $\mathbf{w} = (\bm{w}_{k})_{1 \leq k \leq K}$ and class-wise blending parameters $\bm{\alpha} = (\alpha_k)_{1 \leq k \leq K}$:
\begin{equation}
\label{CE-loss}
L(\mathbf{w}, \bm{\alpha} )= -\frac{1}{N}\sum_{i=1}^{N}\sum_{k=1}^{K} y_{ik}\ln{p_{ik}(\mathbf{w}, \bm{\alpha})}
\end{equation}
where softmax probability outputs $p_{ik}$ are now given by:
\begin{equation}
\label{vision-langage-softmax}
p_{ik}(\mathbf{w}, \bm{\alpha}) =   \frac{\exp \left ( \bm{f}_{i}^{t} (\bm{w}_{k} + \alpha_k \bm{t}_{k}) \right) }{\sum_{j=1}^{K}\exp \left (\bm{f}_{i}^{t} (\bm{w}_{j} + \alpha_j \bm{t}_{j}) \right )} 
\end{equation}
and $\bm{t}_{k}$ are the fixed pre-trained embeddings of the text templates used in zero-shot CLIP \cite{radford2021clip}. Clearly, the objective function defined by (\ref{CE-loss}) and (\ref{vision-langage-softmax}) could be viewed as a generalization of the vision-encoder CE loss used in the standard LP \cite{radford2021clip}. Indeed, the latter corresponds to setting $\alpha_k = 0 \, \forall k$, i.e., no text knowledge.
As we will see in our experimental ablation over different objective functions (Table~\ref{table:ablation-loss-and-optimizers}), introducing the text knowledge ($\alpha_k > 0 \, \forall k$) has a substantial effect on performances. Also, making $\alpha_k$ learnable (rather than fixed) leads to a further significant impact. Indeed, we hypothesize that the optimal blending of the text and visual knowledge is task dependent, which motivates learning it from the context of the support set.

\subsection{Block coordinate Majorize-Minimize descent}

Majorize-Minimize (MM) \cite{Lange2000} is a very general optimization principle, which includes different classes of standard optimizers such as gradient descent, 
concave-convex procedures and expectation-maximization. Let $\mathbf{v} = (\mathbf{w}, \bm{\alpha}) \in {\mathbb R}^{K(D+1)}$ denote the overall vector of 
variables in our case, with $D$ being the dimension of the feature embeddings. At each iteration, the MM procedure updates the variable as the minimum of a
{\em majorizing} function, i.e., an upper bound on the original objective, which is tight at the current iteration $j$: 
$L(\mathbf{v}) \leq M(\mathbf{v},\mathbf{v}^j)$ and $L(\mathbf{v}^j)= M(\mathbf{v}^j, \mathbf{v}^j)$. Thus, update step $\mathbf{v}^{j+1} = \min_{\mathbf{v}} M(\mathbf{v}, \mathbf{v}^j)$ guarantees that the original objective does not increase at each iteration\footnote{This assumes, of course, that minimizing $M(\mathbf{v},\mathbf{v}^j)$ over $\mathbf{v}$ could be solved 
to global optimality and is easier than the original problem.}: $L(\mathbf{v}^{j+1}) \leq M(\mathbf{v}^{j+1},\mathbf{v}^j) \leq M(\mathbf{v}^{j},\mathbf{v}^j) = L(\mathbf{v}^{j})$. 
Therefore, in MM algorithms, step sizes are {\em implicit} in the definition of the majorizing function, unlike standard gradient-descent practices, in which 
the step sizes ({\em a.k.a} learning rates) are intensively searched over validation sets, via running the optimizer several times. 

In this work, we exploit the Lipschitz-gradient continuity of our convex objective in (\ref{CE-loss}), i.e., bounds on the maximum eigen values of the Hessian 
matrices (Prop.~\ref{Lipschitz-constants}), thereby building majorizing functions with data-driven, task-specific step sizes. This removes the need for 
validation searches for the optimization hyper-parameters, reducing the computational load for fine-tuning (Table~\ref{table:ablation-loss-and-optimizers}), while 
yielding performances on par with the best learning rates found with validation (Fig.~\ref{fig:GD}). Also, interestingly, the Lipschitz-based learning rates computed 
from our derivation in Prop.~\ref{Lipschitz-constants} are orders-of-magnitude larger than those typically used in deep learning, yielding steeper decreases towards the minimum. 
Before giving proper majorizing functions for our convex, gradient-Lipschitz function in (\ref{CE-loss}), let us first point to the following results, well-known in convex 
optimization \cite{Bubek2015}. While these results are text-book knowledge in optimization, they enable to connect the general MM principle to gradient descent, motivating 
the data-driven, task-specific step sizes we derive in Prop. \ref{Lipschitz-constants} and the block-coordinate MM optimizer we propose in Alg.~\ref{Algorithm-MM-w-alpha}.     

\begin{lemma}(\cite[p. 268]{Bubek2015})
\label{Descent-lemma}
Assume $L(\mathbf{v})$ is a twice-differentiable function, which has a Lipschitz continuous gradient, i.e., there exists a 
strictly positive Lipschitz constant $\gamma$ such that $\nabla^2 L(\mathbf{v}) \preceq \gamma \mathbf{I}$, with $\mathbf{I}$  
the identity matrix. Then, the following quadratic bound is a majorizing function for $L$ at iteration $j$:
\begin{equation}
\label{quadratic-bound}
 M(\mathbf{v},\mathbf{v}^j) = L(\mathbf{v}^j) + \nabla L(\mathbf{v}^j)^t(\mathbf{v}-\mathbf{v}^j) + \frac{\gamma}{2} \|\mathbf{v}-\mathbf{v}^j \|^2
\end{equation}
Furthermore, a specific gradient step, with learning rate $\frac{1}{\gamma}$ minimizes $M$, i.e., 
$\mathbf{v}^{j+1} = \mathbf{v}^{j} - \frac{1}{\gamma} \nabla L(\mathbf{v}^{j}) = \arg \min_{\mathbf{v}} {M(\mathbf{v},\mathbf{v}^j)}$, and guarantees 
that objective $L$ decreases by at least $\frac{1}{2\gamma} \|\nabla L(\mathbf{v})\|^2$:
\begin{equation}
 L(\mathbf{v}^{j+1}) \leq L(\mathbf{v}^{j}) - \frac{1}{2\gamma} \|\nabla L(\mathbf{v})\|^2  
\end{equation}
\end{lemma}
Moreover, the following Theorem, which follows from Lemma \ref{Descent-lemma}, establishes the {\em sublinear convergence} of the MM procedure 
using bound (\ref{quadratic-bound}), i.e., a convergence rate of $O(1/J)$, $J$ being the total number of iterations.
\begin{theorem}(\cite[p. 267]{Bubek2015}) 
\label{Sublinear-convergence-theorem}
For convex, twice-differentiable function $L(\mathbf{v})$, which has a $\gamma$-Lipschitz gradient, performing 
$J$ updates $\mathbf{v}^{j+1} = \mathbf{v}^{j} - \frac{1}{\gamma} \nabla L(\mathbf{v}^{j})$, starting from initialization $\mathbf{v}^{0}$, will yield a solution that satisfies:
\begin{equation}
\|L(\mathbf{v}^{J}) - L(\mathbf{v}^{*})\| \leq \frac{\gamma}{2J} \|\mathbf{v}^{0} - \mathbf{v}^{*} \|
\end{equation}
where $L(\mathbf{v}^{*})$ is the optimal value. 
\end{theorem}
For completeness, we provide the proofs of these well-known results in the supplemental material. 
Clearly, Lemma \ref{Descent-lemma} and Theorem \ref{Sublinear-convergence-theorem} prescribe a learning rate of $\frac{1}{\gamma}$ for a function that 
has a $\gamma$-Lipschitz gradient. One valid Lipschitz constant would be the maximum eigen value of the Hessian of our objective in (\ref{CE-loss}), 
which provides a majorizing function of the form in Eq.~(\ref{quadratic-bound}) and data-driven learning rates. However, a naive spectral decomposition 
of the Hessian matrices (to obtain the maximum eigen value) could be computationally intensive. For instance, for ImageNet, the Hessian of our objective is 
of size $K (D+1) \times K (D+1) \approx 1M \times 1M$, as $D=1024$ and $K=1000$. In Prop. \ref{Lipschitz-constants}, we derive approximate global and  
block-wise Lipschitz constants that can be computed efficiently (i.e., evaluating the maximum eigen value of a single $D \times D$ matrix).    

\paragraph{Block-coordinate updates} Our procedure provided in Alg. \ref{Algorithm-MM-w-alpha} belongs to the family of Block Majorize-Minimize (BMM) methods, well 
studied in the optimization community \cite{Hong2017}. To minimize a multi-block objective, as in our case where the blocks correspond to variables $\mathbf{w}$ and $\bm{\alpha}$, 
we minimize one or many successive majorizing functions of the objective in each block, with the other block fixed, in a cyclic order:
\begin{equation}
\label{block-w}
L(\mathbf{w}^j, \bm{\alpha}) + \nabla L_{\mathbf{w}}(\mathbf{w}^j)^t(\mathbf{w}-\mathbf{w}^j) + \frac{\gamma_{\mathbf{w}}}{2} \|\mathbf{w}-\mathbf{w}^j \|^2
\end{equation}
\begin{equation}
\label{block-alpha}
L(\mathbf{w}, \bm{\alpha}^j) + \nabla L_{\bm{\alpha}}(\bm{\alpha}^j)^t(\bm{\alpha}-\bm{\alpha}^j) + \frac{\gamma_{\bm{\alpha}}}{2} \|\bm{\alpha}-\bm{\alpha}^j \|^2
\end{equation}
where in (\ref{block-w}), block $\bm{\alpha}$ is fixed and, in (\ref{block-alpha}), $\mathbf{w}$ is fixed. 
$\nabla L_{\mathbf{w}}$ and $\nabla L_{\bm{\alpha}}$ denote block-wise gradients, and $(\gamma_{\mathbf{w}}, \gamma_{\bm{\alpha}})$ are the {\em block Lipschitz constants}.     
Accommodating different choices of the block-cycling strategies and majorizing functions, BMM includes a breadth of optimizers as particular cases, such the well-known block coordinate gradient descent (BCGD) \cite{Beck2013} and its projection-based variant. Indeed, the so-called Gauss-Seidel cycling \cite{Hong2017} alternates steps (\ref{block-w}) and (\ref{block-alpha}), which corresponds to the BCGD method. One could also performs many successive steps in one block, as we do in Alg.~\ref{Algorithm-MM-w-alpha}, which corresponds to the so-called Essentially-Cyclic\footnote{Essentially-Cyclic means that there is a period during which each block is updated at least once.} strategy \cite{Hong2017}. Importantly, for a fairly large spectrum of choices of the cycling strategies, BMM enjoys the same {\em sublinear convergence} property as MM for convex objectives (with each block-wise update decreasing the objective), provided that the majorizing functions are strongly convex; see Theorem 3.1 in \cite{Hong2017}. This is the case for (\ref{block-w}) and (\ref{block-alpha}). In our experiments, we observed that this block-wise variant performs better than a single-block MM; see Table~\ref{table:ablation-loss-and-optimizers}.
This might be explained by the fact that each block of variables has a dedicated step size. 
In the supplemental material, we provide results for different block-cycling strategies.

\paragraph{Global and block-coordinatewise Lipschitz constants}

In the following, we derive approximate global and block-coordinatewise Lipschitz constants for our objective function in (\ref{CE-loss}), which could be evaluated 
efficiently. We deploy these in Alg.~\ref{Algorithm-MM-w-alpha}, to compute data-driven, block-wise learning rates for updating visual prototypes 
$\mathbf{w}$ and blending parameters $\bm{\alpha}$. 

\begin{prop}
\label{Lipschitz-constants}
Considering the blocks of variables $\mathbf{w}$ and $\bm{\alpha}$, the gradient of our objective $L$ in (\ref{CE-loss}) is block-coordinatewise Lipschitz continuous. 
For $\tau_1 \geq 2$, it has the following block Lipschitz constant for the set of variables in $\mathbf{w}$:
\begin{equation}
\label{Lipschitz-constant-w}
\gamma_{\mathbf{w}} = \frac{\tau_1}{4 N} \lambda_{max} \left (\sum_{i=1}^N  (\bm{f}_i \bm{f}_i^t) \right ) 
\end{equation}
where $\lambda_{max}(\bm{A})$ denotes the maximum eigenvalue of matrix $\bm{A}$. Furthermore, for $\tau_1 \geq 1$, the expression in Eq. (\ref{Lipschitz-constant-w}) provides a tighter but approximate block Lipschitz constant. 
Similarly, for $\tau_2 \geq 1$, we have following approximate block Lipschitz constant for the variables in $\bm{\alpha}$:
\begin{equation}
\label{Lipschitz-constant-alpha}
\gamma_{\bm{\alpha}} = \max_k \frac{\tau_2}{4 N}\sum_{i=1}^N (\bm{f}_{i}^t\bm{t}_{k})^2 
\end{equation}
Finally, for $\tau \geq 2$, the following expression provides an approximate global Lipschitz constant for objective (\ref{CE-loss}), i.e., w.r.t all variables $\mathbf{v} = (\mathbf{w}, \bm{\alpha})$: 
\begin{equation}
\label{Global-Lipschitz-constant}
\gamma = \tau \max (\gamma_{\mathbf{w}}, \gamma_{\bm{\alpha}})
\end{equation}
\end{prop}

\begin{proof}
The details are deferred to the supplemental material. The main ingredients of the proof
are based on the Gershgorin circle theorem
and the variational characterization of the maximum eigenvalue,
following the min-max theorem, also referred to as the variational principle. 
\end{proof}

\subsection{Initialization of the variables}

In the following Prop.~\ref{Approx-minimum-CE-v}, we derive approximations of the minima of our objective function (\ref{CE-loss}), which yield a data-informed initialization of the variables.
Indeed, the expressions we obtain in Eqs.~(\ref{hard-mean-1}) and (\ref{hard-mean-alpha}) suggest initial guesses for variables $\mathbf{w}$ and $\bm{\alpha}$. 
Interestingly, and as will be confirmed by our experiments (Table~\ref{table:classifier_init}), such an initialization yields substantially lower values of the minimized loss 
than a random initialization. Furthermore, surprisingly, using this initialization for a {\em training-free} prediction yields better performances than the training-free version of the recent Tip-Adapter-F approach \cite{zhang2022tip}. The details of our training-free version are provided in the supplemental material. 
\begin{prop}
\label{Approx-minimum-CE-v}
The cross-entropy in Eq.~(\ref{CE-loss}) could be written as the sum of two convex functions, i.e., $L = g_1 + g_2$, such that, $\forall k \in [1, \dots, K]$, the
minimum of $g_1$ w.r.t $\bm{w}_{k}$ is co-linear to the hard mean vector of features within class $k$:
\begin{equation}
\label{hard-mean-1}
\argmin_{\bm{w}_{k}} g_1 = \frac{1}{\lambda N}\sum_{i=1}^N y_{ik}\bm{f}_{i} \propto \frac{\sum_{i=1}^{N} y_{ik}\bm{f}_{i}}{\sum_{i=1}^{N} y_{ik}} 
\end{equation}
and the minimum of $g_2$ w.r.t $\bm{w}_{k}$ is co-linear to the soft mean vector of features within class $k$: 
\begin{equation}
\label{soft-mean-1}
\argmin_{\bm{w}_{k}} g_2  = \frac{1}{\lambda N}\sum_{i=1}^N p_{ik}\bm{f}_{i} \propto \frac{\sum_{i=1}^{N} p_{ik}\bm{f}_{i}}{\sum_{i=1}^{N} p_{ik}}  
\end{equation}
where $\lambda \leq \min_k \lambda_{\textsuperscript{min}}(\bm{A}_{k})$, $\bm{A}_{k} = \frac{1}{N}\sum_{i=1}^N (p_{ik} - p_{ik}^2) \bm{f}_i \bm{f}_i^t$ and
$\lambda_{\textsuperscript{min}} (\bm{A})$ denotes the smallest eigenvalue of matrix $\bm{A}$. 
\end{prop}
\begin{proof}
We defer the details, including the full expressions of convex functions $g_1$ and $g_2$, to the supplemental material. 
\end{proof}

Similarly to the development in Prop.~\ref{Approx-minimum-CE-v}, one could decompose (\ref{CE-loss}) as the sum of two convex functions, i.e., $L = h_1 + h_2$, such 
that, $\forall k \in [1, \dots, K]$, the minima of $h_1$ and $h_2$ w.r.t $\alpha_{k}$ could be written, up to a multiplicative positive factor, as the hard and soft means of the cosine similarities between the image and text embeddings:
\begin{equation}
\label{hard-mean-alpha}
\argmin_{\alpha_{k}} h_1 = \frac{1}{\beta N} \sum_{i=1}^{N} y_{ik}\bm{f}_{i}^t\bm{t}_{k}
\end{equation}
\begin{equation}
\label{soft-mean-alpha}
\argmin_{\alpha_{k}} h_2 = \frac{1}{\beta N} \sum_{i=1}^{N} p_{ik}\bm{f}_{i}^t\bm{t}_{k}
\end{equation}
Here, $\beta= \min_k \frac{1}{N}\sum_{i=1}^N (p_{ik} - p_{ik}^2) (\bm{f}_{i}^t\bm{t}_{k})^2$, and $h_1$ (respectively $h_2$) has the same expression as $g_2$ (respectively $g_1$), except that term $\frac{\lambda}{2}\sum_{k=1}^{K} \|\bm{w}_{k} \|^2$ is replaced by $\frac{\beta}{2}\|\bm{\alpha}\|^2$; see the supplemental material for the expressions of $g_1$ and $g_2$. 

\begin{center}
\RestyleAlgo{ruled}
	\begin{algorithm}
 iter$_{\mathbf{w}}$ = 10; iter$_{\bm{\alpha}}$ = 1; $\tau_1=1$; $\tau_2=16$; $\lambda=\frac{1}{N}$; $\beta=\frac{1}{250 K}$  \\
Initialize ${\mathbf w}^{0,0}$ {\color{blue}\tcp{\small Using (\ref{hard-mean-1}) for each $k$}} 
Initialize $\bm{\alpha}^{0,0}$  {\color{blue}\tcp{\small Using (\ref{hard-mean-alpha}) for each $k$}} 
\For{$j = 0, 1, \ldots$}
{ \For{$l_{1} = 0, 1, \ldots,$ {\em iter}$_{\mathbf{w}}$}
{
$\mathbf{w}^{j,l_{1}+1} = \mathbf{w}^{j,l_{1}} - \frac{1}{\gamma_{{\mathbf w}}} \nabla L_{\mathbf{w}}(\mathbf{w}^{j,l_{1}},\bm{\alpha}^{j,0})$ \\ {\color{blue}\tcp{\small $\frac{1}{\gamma_{\mathbf{w}}}$ from Eq.(\ref{Lipschitz-constant-w})}}} 
$\mathbf{w}^{j+1,0} = \mathbf{w}^{j,\mbox{{\small iter}}_{\mathbf{w}}}$
\\ 
\For{$l_{2} = 0, 1, \ldots,$ {\em iter}$_{\bm{\alpha}}$}
{
$\bm{\alpha}^{j,l_{2}+1} = \bm{\alpha}^{j,l_{2}} - \frac{1}{\gamma_{\bm{\alpha}}} \nabla L_{\bm{\alpha}} (\mathbf{w}^{j+1,0},\bm{\alpha}^{j,l_{2}})$\\
{\color{blue}\tcp{\small $\frac{1}{\gamma_{\bm{\alpha}}}$ from Eq.(\ref{Lipschitz-constant-alpha})}}
}$\bm{\alpha}^{j+1,0} = \bm{\alpha}^{j,\mbox{{\small iter}}_{\bm{\alpha}}}$
\\ }
\caption{Block coordinate MM (${\mathbf w}$, $\bm{\alpha}$) \label{Algorithm-MM-w-alpha}}
\end{algorithm}
\end{center}

\section{Experiments}
\label{Experiments}

\begin{table*}[ht]
\scriptsize
\renewcommand{\arraystretch}{1.2}
\centering
\begin{tabular}{llcccccc}
\hline
\textbf{Number of shots ($S$)} & & \textbf{1} & \textbf{2} & \textbf{4} & \textbf{8} & \textbf{16}\\
\midrule
& Zero-shot CLIP$_\text{ ICML'21}$\cite{radford2021clip} & \multicolumn{5}{c}{58.89}\\
\midrule
\multirow{4}{*}{\textit{Prompt-Learning}} & CoOp$_\text{ IJCV'22}$\cite{Zhou2022coop} & $59.62 \pm 3.11$ & $63.80 \pm 2.32$  &  $67.23 \pm 1.64$  & $71.30 \pm 0.86$  &  $74.06 \pm 0.55$ \\
& PLOT$_\text{ ICLR'23}$\cite{chen2023plot} & $61.51 \pm 2.91$ &  $65.67 \pm 2.06$ &  $68.39 \pm 1.17$  &  $71.96  \pm 0.70$ & $74.35 \pm 0.66$  \\
& KgCoOp$_\text{ CVPR'23}$\cite{Yao2023kgcoop} & $61.36 \pm 3.04$ & $63.23 \pm 2.06$ & $65.73 \pm 1.15$ &$67.50 \pm 1.11$ & $69.01 \pm 0.79$ \\
& ProGrad$_\text{ ICCV'23}$\cite{Zhu2023prograd} & $62.46 \pm 1.89$ & $65.88 \pm 1.46$ & $68.52 \pm 1.15$ & $71.82 \pm 0.11$ & $73.95 \pm 0.68$\\
\midrule
\multirow{3}{*}{\textit{CLIP-based Adapters}} & CLIP-Adapter$_\text{ IJCV'23}$\cite{Gao2023clipadapter} & $60.32 \pm 0.80 $ & $61.93 \pm 0.93$ & $65.12 \pm 0.80$ & $69.20 \pm 0.56$ & $ 72.57 \pm 0.54$ \\
& Tip-Adapter-F$_\text{ ECCV'22}$\cite{zhang2022tip} & $61.29 \pm0.92$ & $62.94 \pm0.75$ & $66.02 \pm0.80$  & $69.88\pm 0.51$  & $73.82\pm 0.55$ \\
& Tip-Adapter-F*$_\text{ ECCV'22}$\cite{zhang2022tip} & $63.06 \pm1.05$ & $\textbf{66.47} \pm0.65$ & $68.71 \pm0.96$ & $71.78\pm 1.00$ & $74.37\pm 0.35$ \\
\midrule
\multirow{2}{*}{\textit{Linear-Probing}} & Standard LP$_\text{ ICML'21}$\cite{radford2021clip} & $36.10 \pm 1.43$  &  $46.99 \pm 1.29$  &  $56.72 \pm1.20$  &  $64.66 \pm0.55$  & $70.56\pm 0.44$  \\
& \cellcolor{blue!15}{LP++} & \cellcolor{blue!15}{$\textbf{63.43} \pm 0.90 $} & \cellcolor{blue!15}{$66.20 \pm 0.72 $} & \cellcolor{blue!15}{$\textbf{69.16} \pm 0.79 $} & \cellcolor{blue!15}{$\textbf{72.04} \pm 0.46$ } & \cellcolor{blue!15}{$\textbf{74.42} \pm 0.45 $}\\
\bottomrule
\end{tabular}
\caption{\textbf{Comparison to state-of-the-art methods.} Average classification accuracy (\%) on 11 benchmarks, 
with standard derivation over 10 sampled support sets for each dataset. The best values are highlighted in bold.}
\label{table:resultOther_AllDatasets}
\end{table*}

\subsection{Datasets and Implementation details}

Following the CLIP-based few-shot adaptation literature \cite{Yao2023kgcoop, zhang2022tip}, we conduct the main experiments on 11 public classification data sets: Caltech101 \cite{caltech101}, ImageNet \cite{imagenet}, DTD \cite{dtd}, OxfordPets \cite{oxfordpets}, Flowers102 \cite{flowers102}, StandfordCars \cite{stanfordcars}, Food101 \cite{food101}, FGVCAircraft \cite{fgvcaircraft}, SUN397 \cite{sun397}, EuroSAT \cite{eurosat} and UCF101 \cite{ucf101}. We follow standard practices \cite{radford2021clip} and consider $S=\{1, 2, 4, 8, 16\}$ shots 
for model adaptation, which are randomly sampled for each data set. 

\noindent \textbf{Towards a fair validation set}. Apparently, prior works on this problem \cite{zhang2022tip} have resorted to a large set of validation samples to adjust their hyper-parameters. For the sake of fairness, 
we tune the hyper-parameters across all the methods based on a small validation set, which contains as many samples (i.e., shots) as the training set. 
Furthermore, to avoid the potential overfitting on the few training samples, 
we adopt an early stopping strategy 
on this validation set. 

\noindent \textbf{General Setting}. 
While the existing works evaluate methods based on either a single or three random tasks (support sets) \cite{zhang2022tip, Zhou2022coop}, we found that, for some datasets, the chosen support samples may not be representative of the class, leading to large standard deviations in low-shot scenarios (see ~\cref{fig:avg-std}). To ensure fair comparisons, we evaluate all the methods by averaging their classification accuracies over 10 randomly 
sampled tasks, for each dataset. In all the experiments, we employ ResNet-50 \cite{he2015deep} as the visual encoder for the CLIP backbone. {\em It is important to note that, for our BMM procedure in Alg. \ref{Algorithm-MM-w-alpha}, the optimizer hyper-parameters remain fixed across all the datasets}. We use the validation set only to find the best model, via a single run of our BMM procedure with a fixed number of variable updates, i.e., 300 gradient updates including all the blocks of variables. 

\subsection{Baselines}
\label{sec:baselines}
We benchmark the proposed LP++ against relevant state-of-the-art methods in the few-shot adaptation of CLIP-based models. 
We first resort to zero-shot CLIP as the standard baseline, which only leverages the knowledge learned by the pre-trained CLIP model. 
Also, we includes the standard LP baseline, whose implementation is done following \cite{Zhou2022coop,radford2021clip}. More concretely, this baseline optimizes the standard cross-entropy loss, which corresponds to $\alpha_k > 0 \, \forall k$ in our generalization in (\ref{CE-loss}), using the L-BFGS \cite{nocedal1980updating} optimizer\footnote{L-BFGS (Limited-memory BFGS) aims to find the minimum of objective function using the second order method. It estimates the Hessian matrix based on recent gradients only, enabling it to determine the steepest direction for achieving the optimal solution. Additionally, it is implemented with line searches to automatically determine the optimal step size.}. It also includes an $l_2$-regularizer, whose balancing weight is set based on the validation set. 


\noindent \textbf{CLIP-based adapters.}
We benchmark LP++ 
against two popular adapter-based few-shot 
approaches: CLIP-adapter \cite{Gao2023clipadapter} and TIP-adapter ~\cite{zhang2022tip}. As exposed earlier, several popular works follow unfair practices by resorting to a larger validation set, or even to the entire test set, to adjust their key hyperparameters --as well as their model selection criteria (i.e., epochs)-- for each task. 
For the sake of fairness, we re-implement Tip-Adapter-F and report the results in 2 different settings. In the first setting (Tip-Adapter-F), we set the two crucial hyper-parameters of this method to 1, keeping them fixed during training, and adopt early stopping based on the validation set. 
In the second setting, referred to as Tip-Adapter-F$^*$, 
we perform intensive grid-search on the validation set to find the best values for these hyperparameters at initialization, which incurs an additional time complexity burden compared to Tip-Adapter-F. 

\noindent \textbf{Prompt learning.} We further compare the proposed LP++ to relevant prompt-learning methods, including CoOp\cite{Zhou2022coop} and more recent variants, such as PLOT\cite{chen2023plot}, KgCoOp\cite{Yao2023kgcoop} and ProGrad\cite{Zhu2023prograd}. We also apply early stopping here based on the performances on the validation set. 


\begin{figure*}
     \centering

     \includegraphics[width=0.85\linewidth]{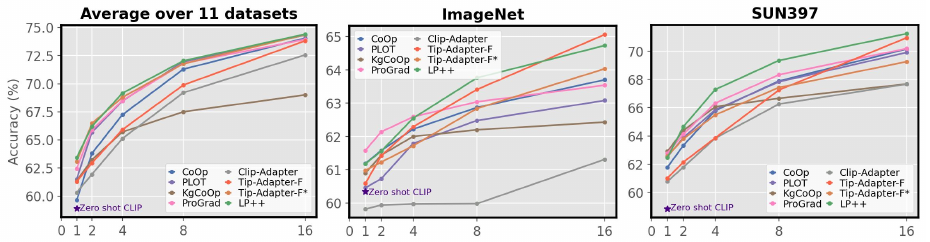}

    \caption{Quantitative performance of different adaptation methods 
    on the 11 benchmarks (mean), as well as in two other datasets, averaged over 10 tasks (additional figures on the remaining 9 datasets can be found in Appendix,  ~\cref{sec:additional_results}). }
    \label{fig:clip_mag}
\end{figure*}

\subsection{Results}


\paragraph{Comparison to the state-of-the-art}
In Table~\ref{table:resultOther_AllDatasets}, we present the quantitative results obtained by LP++ (Alg. \ref{Algorithm-MM-w-alpha}) and the relevant literature in the task of efficient adaptation of VLMs. We report the average classification accuracy and standard deviation, across 11 classification benchmarks. From these results, one can make several observations. First, while the standard LP baseline largely under-performs the existing adaptation methods, our improved version, LP++, brings significant performance gains, particularly in the low-labeled data regimes. It is important to stress that the standard LP baseline integrates only the visual features extracted from CLIP, disregarding the text-encoder knowledge. This contrasts with the existing adapters, which leverage both image and text information. Therefore, these results evidence that \textbf{the potential of LP has been severely underestimated in the existing literature.} Second, when the model selection process is performed fairly, i.e., using a small validation set, their performances fall behind the proposed method (from 1\% to 5\%), despite being arguably more complex approaches. In particular, among the adapter-based strategies, only Tip-Adapter-F* yields performances on par with LP++, but at the cost of increasing computational load, due to an additional intensive grid-search over its hyper-parameters. If we look at prompt-learning methods, 
ProGrad and PLOT might be considered as competitors of LP++, particularly as the number of shots increases. Nevertheless, as already discussed, these methods are computationally inefficient compared to adapters, 
and do not enable black-box adaptation. 
The overall performance across all the methods 
is depicted in ~\cref{fig:clip_mag} for the average over the 11 benchmarks, as well as in two datasets, showing that LP++ typically outperforms existing methods under different few-shot scenarios. A more detailed analysis is deferred to the Appendix, Sec. \ref{sec:additional_results}.

\begin{figure*}[ht!]
\centering
\includegraphics[width=0.85\linewidth]{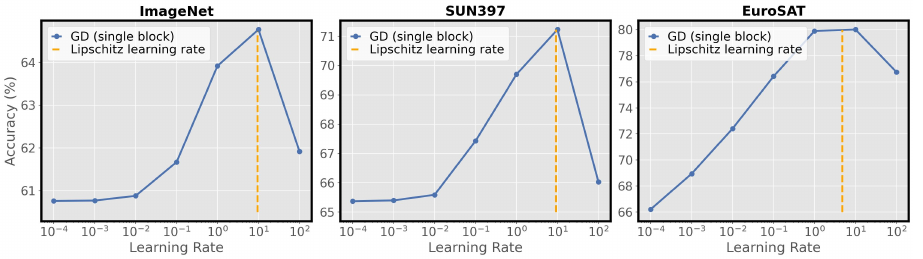}
\caption{Single-block GD performance as a function of different values of the learning rates. The dotted vertical line shows the Lipschitz-based, data-driven learning rate.}
\label{fig:GD}
\end{figure*}



\noindent \textbf{Ablation on the loss functions and different optimization strategies.}
\Cref{table:ablation-loss-and-optimizers} reports the test accuracy and run time for different optimizers and loss functions, including the standard CE loss ($\alpha_k = 0 \, \forall k$) and our loss with learnable blending parameters. Additionally, for LP++, we evaluate our loss with fixed blending parameters ($\alpha_k = 1 \, \forall k$). Independently of the optimizer used, the main takeway from \Cref{table:ablation-loss-and-optimizers} is that introducing the text knowledge and making $\alpha_k$ learnable (rather than fixed) have a substantial impact on accuracy. As for evaluating the optimizers, and for a fair comparison, we use a fixed budget for the number of variable updates for all the optimizers (i.e., 300 updates). In the case of LP++, this corresponds to the total number of updates for all the blocks. Also, for all optimizers, we initialize $\mathbf{w}$ and $\bm{\alpha}$ following Eqs. (\ref{hard-mean-1}) and (\ref{hard-mean-alpha}). 
We first consider two popular optimizers, i.e., GD and ADAM, and deploy them in two different settings. First, we run each optimizer 7 times, with each run corresponding to a learning rate in the range $[10^{-4}, 10^{2}]$. Then, we record the best performances obtained on the validation set; see GD (optimum) and ADAM (optimum) in \Cref{table:ablation-loss-and-optimizers}. This follows the standard practices in deep learning, i.e., searching for the learning rates over validation sets, which incurs additional computation overhead; see the time column in \Cref{table:ablation-loss-and-optimizers}. Second, we run GD and ADAM with our data-driven learning rate, as prescribed by the approximate Lipschitz constant we derived 
in Eq.~(\ref{Global-Lipschitz-constant}), with $\tau=1$; see GD (our Lipschitz cst) and ADAM (our Lipschitz cst) in \Cref{table:ablation-loss-and-optimizers}. Note that, in this case, GD 
corresponds to LP++ with a single block.    
Furthermore, we include the L-BFGS in these comparisons, with its initial learning rate set to 1. 
For L-BFGS, implementing a line search for the optimal step size also introduces an additional computational overhead. 
As highlighted by \Cref{table:ablation-loss-and-optimizers}, our method removes the need for validation searches for the optimization hyper-parameters, thanks to its 
data-driven, task-specific step sizes, thereby reducing the computational load for fine-tuning. In the meanwhile, it yields performances on par with those obtained with the best learning rates found with the validation set.
In Fig.~\ref{fig:GD}, we plot the performances of single-block GD vs. the learning rates in the range $[10^{-4}, 10^{2}]$, for three datasets. 
We observe that our Lipschitz-based, task-specific step sizes match the optimal ones found on the validation set, although these vary among the datasets. 
Also, interestingly, these Lipschitz-based step sizes are orders-of-magnitude larger than those used in deep learning, which are, typically, within interval $[10^{-4}, 10^{-2}]$; see \cite{zhang2022tip}, for instance.

\begin{table}[h]
\scriptsize
\setlength{\tabcolsep}{4pt}
\centering
\begin{tabular}{lccccc}
\toprule
 Optimization Method & \multicolumn{2}{c}{Standard loss ($\alpha_k=0$)}  & \multicolumn{2}{c}{Our loss} & Time\\
\hline
\midrule
& 1 shot & 16 shot & 1 shot & 16 shot&\\
\midrule
{LP++ ($\alpha_{k}$ is learnt)}& {35.55} & {69.75}  & \textbf{63.43}  & \underline{74.42} & 0.78s\\
LP++ ($\alpha_{k}=1$)& -    & -   &  46.37 & 69.41&  0.72s\\
 \hline
GD (our Lipschitz cst) &  35.55 & 69.75   & 63.04	 & 74.28& 0.86s \\
GD (optimum)  & 35.64  & 69.67    & 62.93  & \textbf{74.55} & 6.02s\\
\hline
ADAM (our Lipschitz cst) &  26.22	  & 64.53  &  25.44	  & 64.62  &  0.98s\\
ADAM (optimum) & 35.90  &  69.73   &  \underline{63.07} & 74.23 & 6.89s\\
\hline
L-BFGS \cite{nocedal1980updating} & 34.54   & 67.44 &  62.09 & 72.82 & 6.73s\\
\bottomrule
\end{tabular}
\label{optimizer}
\caption{Accuracy and run time for different optimizers and loss functions (average over 11 datasets). The running time is recorded for our loss (16 shots), and is averaged over 11 datasets. The best result is marked in bold, and the second best is underlined.}
\label{table:ablation-loss-and-optimizers}
\end{table}

\noindent \textbf{How to initialize the classifier?}
To justify empirically the advantages brought by initializing the classifier weights following Eq. (\ref{hard-mean-1}), and the blending parameters in Eq. (\ref{hard-mean-alpha}), we evaluate objective (\ref{CE-loss}) at the beginning of the training, as well as the training-free test accuracy of three methods: random initialization, and the training-free version of both the Tip-adapter-F method \cite{zhang2022tip} (i.e., Tip-adapter) and our method (we provide more details on training-free LP++ in the appendix). 
The results in Table~\ref{table:classifier_init} confirm empirically the technical observations in Eqs. (\ref{hard-mean-1}) and (\ref{hard-mean-alpha}), which prescribe initial guesses for our problem's variables.


\begin{table}[h]
\scriptsize
\centering
\begin{tabular}{cccccc}
\hline
\textbf{Number of shots ($S$)} & \textbf{1} & \textbf{2} & \textbf{4} & \textbf{8} & \textbf{16}\\
\hline
\midrule
Models & \multicolumn{5}{c}{Initial Loss ($L_0$)}\\
\hdashline
Random Initialization & 21.45 & 21.42 & 21.31 & 21.27 & 21.32  \\
Proposed Initialization & 1.60 & 1.54 & 1.47 & 1.35 & 1.21  \\
\midrule
Models & \multicolumn{5}{c}{Initial test accuracy ($\textrm{Acc}_0$)}\\
\hdashline
Random Initialization & 17.79 & 17.92 & 18.03 & 18.10 & 18.22  \\
Tip-adapter\cite{zhang2022tip} & 59.28 & 59.72 & 60.55 & 62.09 & 64.29  \\
Proposed Initialization & 59.70 & 60.66 & 62.04 & 64.16 & 66.20  \\
\bottomrule
\end{tabular}
\caption{Comparison of the initial loss and 
test accuracy in the \textit{training-free} scenario: random initialization, Tip-adapter and the proposed initialization. The results are averaged over 11 datasets.}
\label{table:classifier_init}
\end{table}

\noindent \textbf{Computational overhead.} As the literature on adapting VLMs is gaining popularity, it is essential to evaluate the extent to which novel methods are efficient. 
To do this, we report the overall computational overhead of the 
approaches studied in this work, which includes the time required for training and for finding the hyper-parameters, when applicable. We also indicate whether these methods enable black-box adaptation which, in our perspective, is a critical aspect in novel strategies aiming to address practical, real-world demands. The numerical values in \Cref{table:time_Imagenet} show that, in addition to yielding state-of-the-art performance (shown in previous sections), LP++ is the most efficient method (by several orders of magnitude), and does not require to access the internal representations
of the pre-trained models.


\begin{table}[h]
\footnotesize
\setlength{\tabcolsep}{4pt}
\centering
\begin{tabular}{lccc}
\toprule
 Methods & Overall Time  & BlackBox & \# Parameters\\
\midrule 
\midrule
CoOp\cite{Zhou2022coop}  & $\sim$ 17h &  \xmark & $K\times M\times D$\\
PLOT$-2$\cite{chen2023plot} & $\sim$ 10h& \xmark & $P\times K\times M\times D$\\
KgCoOp\cite{Yao2023kgcoop}  & $\sim$ 4h& \xmark &   $K\times M\times D$\\
ProGrad\cite{Zhu2023prograd}  & $\sim$ 20h & \xmark & $K\times M\times D$\\
\midrule
 Clip-Adapter\cite{Gao2023clipadapter} & $\sim$ 40min & $\checkmark$ & $2(D_{1}\times D)$ \\
Tip-adapter-F\cite{zhang2022tip} & $\sim$ 6min & $\checkmark$ & $K\times S\times D$\\
Tip-adapter-F*\cite{zhang2022tip}  & $\sim$ 50min & $\checkmark$ &$K\times S\times D$ \\
\midrule
Standard LP\cite{radford2021clip}  & 3min  & $\checkmark$ & $K\times D$\\
LP++  & $\sim$ 2s &$\checkmark$ & $K(D+1)$\\
\bottomrule
\end{tabular}
\caption{Run time and suitability to black-box scenarios for different methods on 16-shot ImageNet. All the experiments are performed on a single NVIDIA RTX A6000 GPU, except for PLOT$-2$, which is evaluated on two A6000 GPUs. $D_{1}=256$, and $D=1024$. The number of context tokens $M$ is set to 16. For PLOT, $P=4$ is the number of prompts.}
\label{table:time_Imagenet}
\end{table}

\section{Conclusion}
\label{Conclusion}
We introduced LP++, a strong linear probe for few-shot CLIP adaptation. A specific modeling of the classifier weights, blending visual prototypes and text embeddings via learnable multipliers, along with convex-optimization ingredients, often overlooked in deep learning practices, led to the surprising results. While
the findings of this work do not invalidate the promise of prompt learning and adaptation research, we believe LP++ could be used as a baseline to measure progress in these strongly emergent areas.

{
    \small
    \bibliographystyle{ieeenat_fullname}
    \bibliography{main}
}


\clearpage
\setcounter{page}{1}
\maketitlesupplementary

\section{Proof of Prop. \ref{Lipschitz-constants}}

Let us start with $H_{{\mathbf w}}$, the Hessian matrix of objective function (\ref{CE-loss}) with respect to block ${\mathbf w} \in {\mathbf R}^{KD}$, with block $\bm{\alpha}$ being fixed. 
Let us introduce vector $\bm{F}_{ik} \in \mathbb{R}^{K D}$, which takes the following form:
\begin{equation}
\bm{F}_{ik} = [0, \dots, 0, \bm{f}_{i}^{t}, 0, \dots, 0]^{t}
\end{equation}
where embedding vector $\bm{f}_i \in \mathbb{R}^D$ corresponds to the elements of $\bm{F}_{ik}$ going from the $((k-1)D+1)^{\mbox{{\small th}}}$ to the $(kD)^{\mbox{{\small th}}}$ position, with the rest of elements of $\bm{F}_{ik}$ all being equal to zero.  
Block-coordinatewise $KD \times KD$ Hessian matrix $H_{{\mathbf w}}$ reads:
\begin{align}
\label{appendix-Hessian-w}
H_{{\mathbf w}} & = \frac{1}{N} \sum_{i=1}^N \underbrace{\sum_{k=1}^K  p_{ik} \bm{F}_{ik} \bm{F}_{ik}^t}_{\bm{D}_i \otimes (\bm{f}_i \bm{f}_i^t)} \nonumber \\ 
&- \frac{1}{N} \sum_{i=1}^N \underbrace{\sum_{k=1}^K \sum_{k'=1}^K  p_{ik} p_{ik'} \bm{F}_{ik} \bm{F}_{ik'}^t}_{({\mathbf p}_i {\mathbf p}_i^t) \otimes (\bm{f}_i \bm{f}_i^t)} \nonumber \\
& = \frac{1}{N} \sum_{i=1}^N (\bm{D}_i - {\mathbf p}_i {\mathbf p}_i^t) \otimes (\bm{f}_i \bm{f}_i^t)   \nonumber \\   
\end{align}
where $\otimes$ denotes the Kronecker matrix product, $\bm{D}_i$ is the diagonal $K \times K$ matrix whose elements are given by $p_{ik}$:
\[ \bm{D}_i = \mbox{Diag}((p_{ik})_{1 \leq k \leq  K})\]
and ${\mathbf p}_i \in \mathbb{R}^K$ is the vector whose elements are $p_{ik}$:
\[ {\mathbf p}_i = (p_{ik})_{1 \leq k \leq  K}\]
Now, let us show the following intermediate result on diagonally dominant, symmetric matrices $(\bm{D}_i - {\mathbf p}_i {\mathbf p}_i^t)$, which we could establish using the the 
well-known Gershgorin circle theorem.
\begin{lemma}
\label{appendix-max-eigen-value-D}
Let $\lambda$ denotes an eigen value of matrix $(\bm{D}_i - {\mathbf p}_i {\mathbf p}_i^t)$, then $\lambda \in [0, \frac{1}{2}]$. Therefore, $(\bm{D}_i - {\mathbf p}_i {\mathbf p}_i^t)$ is positive semidefinite and its maximum eigen value verifies: 
\[\lambda_{max}(\bm{D}_i - {\mathbf p}_i {\mathbf p}_i^t) \leq \frac{1}{2} \]  
\end{lemma}
\begin{proof}
For $j \in {1, \dots, K}$, let $R_j$ denotes the sum of the absolute values of the non-diagonal entries in the $j^{\mbox{{\small th}}}$ row of a $K \times K$ matrix. 
For $\bm{D}_i - {\mathbf p}_i {\mathbf p}_i^t$, observe the following, due to probability simplex constraint $\sum_{k=1}^K p_{ik} = 1$:
\[R_j = p_{ij} \sum_{k \neq j} p_{ik} = p_{ij} (1-p_{ij}) = p_{ij}- p_{ij}^2 \]   
Also, notice that $R_j$ is equal to the diagonal element of the $j^{\mbox{{\small th}}}$ row, which means matrix $\bm{D}_i - {\mathbf p}_i {\mathbf p}_i^t$ is diagonally dominant.  
The Gershgorin circle theorem states that, for any eigen value $\lambda$ of a matrix, there exists at least a row $j$, so that $\lambda$ is within the disc centered at the diagonal
element of the row and whose radius is $R_j$.
Using this fact for matrix $\bm{D}_i - {\mathbf p}_i {\mathbf p}_i^t$ means that, for any eigen $\lambda$ of the matrix, there exists at least 
one raw $j$ verifying:
\[|\lambda - (p_{ij}- p_{ij}^2)| \leq R_j\]
This yields:
\[ 0 \leq \lambda \leq 2 (p_{ij}- p_{ij}^2) \leq \frac{1}{2}  \] 
The last inequality above is due to $0 \leq p_{ij} - p_{ij}^2 \leq \frac{1}{4}$ for $p_{ij} \in [0, 1]$. Since all the eigen values of matrix $\bm{D}_i - {\mathbf p}_i {\mathbf p}_i^t$ are in $[0, \frac{1}{2}]$, then the maximum eigen value is also within this interval.   
\end{proof}
From Lemma \ref{appendix-max-eigen-value-D}, it it straightforward to see that matrix $\bm{D}_i - {\mathbf p}_i {\mathbf p}_i^t - \frac{1}{2} \bm{I}_K$ is negative semidefinite and, therefore, its Kronecker multiplication by $\bm{f}_i \bm{f}_i^t$ is also negative semidefinite (as $\bm{f}_i \bm{f}_i^t$ is positive semidefinite).   
Hence, the following matrix is negative semi-definite, as it is the sum of negative semidefinite matrices:
\begin{equation}
H_{{\mathbf w}} - \frac{1}{N} \sum_{i=1}^N \frac{1}{2} \bm{I}_K \otimes (\bm{f}_i \bm{f}_i^t)   
\end{equation}
Therefore:
\begin{equation}
\bm{x}^t H_{{\mathbf w}} \bm{x} \leq \bm{x}^t \left ( \frac{1}{N} \sum_{i=1}^N \frac{1}{2} \bm{I}_K \otimes (\bm{f}_i \bm{f}_i^t) \right )  \bm{x} \quad \forall  \bm{x}
\end{equation}
This yields:
\begin{equation}
\label{variational-H-w}
\max_{\|\bm{x}\|=1} \bm{x}^t H_{{\mathbf w}} \bm{x} \leq \max_{\|\bm{x}\|=1} \bm{x}^t \left ( \frac{1}{N} \sum_{i=1}^N \frac{1}{2} \bm{I}_K \otimes (\bm{f}_i \bm{f}_i^t) \right )  \bm{x} 
\end{equation}
Now, recall the following variational characterization of the maximum eigenvalue of a matrix, following the min-max theorem, also referred to as the variational principle: 
\[\lambda_{max}(H) = \max_{\|\bm{x}\|=1} \bm{x}^t H \bm{x} \]
Using this variational characterization and Eq. (\ref{variational-H-w}), we obtain:
\begin{equation}
\lambda_{max} (H_{{\mathbf w}}) \leq \frac{1}{2N} \lambda_{max} \left (\sum_{i=1}^N \bm{I}_K \otimes (\bm{f}_i \bm{f}_i^t) \right )
\end{equation}
Observe that $KD \times KD$ matrix $\sum_{i=1}^N \bm{I}_K \otimes (\bm{f}_i \bm{f}_i^t)$ is block-diagonal with each of its $D \times D$  diagonal blocks corresponding to matrix $\sum_{i=1}^N  \bm{f}_i \bm{f}_i^t$. Therefore, we have:
\[\lambda_{max} \left ( \sum_{i=1}^N \bm{I}_K \otimes (\bm{f}_i \bm{f}_i^t) \right) =  \lambda_{max} \left (\sum_{i=1}^N  \bm{f}_i \bm{f}_i^t \right ) \] 
This means that the expression in Eq. (\ref{Lipschitz-constant-w}) in the paper is an upper bound on the maximum eigen value of $H_{{\mathbf w}}$, for $\tau_1 \geq 2$. Hence, it is a valid block Lipschitz constant for the set of variables in ${\mathbf w}$. 

It is also possible to have a tighter, but {\em approximate} Lipschitz constant by omitting the off-diagonal elements 
in matrices $\bm{D}_i - {\mathbf p}_i {\mathbf p}_i^t$ in the Hessian-matrix expression in Eq. (\ref{appendix-Hessian-w}).
The approximation is motivated by the fact that these matrices are diagonally dominant. This gives the following approximation of 
the Hessian:
\begin{align}
\label{appendix-Hessian-w-approximate}
\tilde{H}_{{\mathbf w}}  = \frac{1}{N} \sum_{i=1}^N \tilde{\bm{D}}_i \otimes (\bm{f}_i \bm{f}_i^t)   \nonumber \\   
\end{align}
where $\tilde{\bm{D}}_i$ is the diagonal $K \times K$ matrix whose elements are given by $p_{ik} - p_{ik}^2, k=1, \dots, K$. Since $p_{ik} - p_{ik}^2 \leq \frac{1}{4}$, we have:
$\lambda_{max}(\tilde{\bm{D}}_i) \leq \frac{1}{4}$. Therefore, following the same arguments as before, we have: 
\begin{equation}
\lambda_{max} (\tilde{H}_{{\mathbf w}}) \leq \frac{1}{4N} \lambda_{max} \left (\sum_{i=1}^N (\bm{f}_i \bm{f}_i^t) \right )
\end{equation}
Hence, $\tau_1 \geq 1$ in Eq. (\ref{Lipschitz-constant-w}) in the paper provides an approximate Lipschitz constant for the block of variables in ${\mathbf w}$.  

Notice that we still need an eigenvalue decomposition of $\sum_{i=1}^N  \bm{f}_i \bm{f}_i^t$. However, for feature embeddings of size $1024$, i.e., a 
$1024 \times 1024$ matrix, this computation, which has to be performed only once, is manageable; it took $0.3$ 
seconds on a single NVIDIA RTX A6000 GPU.

Now, let us look at the Hessian matrix of objective function (\ref{CE-loss}) w.r.t block $\bm{\alpha} \in {\mathbf R}^K$, with the variables in ${\mathbf w}$ being fixed. 
Let vector $\bm{T}_{ik} \in \mathbb{R}^{K}$ takes the form:
\[\bm{T}_{ik} = [0, \dots, 0, \bm{f}_{i}^t\bm{t}_{k}, 0, \dots, 0]^t\]  
where the $k^{\mbox{{\small th}}}$ element is equal to $\bm{f}_{i}^t\bm{t}_{k}$ and the rest of elements are equal to zero.
This block-coordinatewise $K \times K$ Hessian reads:
\begin{align}
\label{H-alpha-appendix}
H_{\bm{\alpha}} & = \frac{1}{N} \sum_{i=1}^N \sum_{k=1}^K p_{ik} \bm{T}_{ik} \bm{T}_{ik}^t
\nonumber \\ 
&- \frac{1}{N} \sum_{i=1}^N \sum_{k=1}^K \sum_{k'=1}^K  p_{ik} p_{ik'} \bm{T}_{ik} \bm{T}_{ik'}^t
\nonumber \\ 
&=  \frac{1}{N}\sum_{i=1}^N (\bm{D}_i - {\mathbf p}_i {\mathbf p}_i^t)  \odot  \bm{T}_{i}  
\end{align}
where $\odot$ denotes the Hadamard product and $\bm{T}_{i}$ is the $K \times K$ matrix whose element at $(k, k') \in [1, \dots, K]^2$ is given by: $\bm{f}_{i}^t\bm{t}_{k}$$\bm{f}_{i}^t\bm{t}_{k'}$. As before, omitting the off-diagonal elements in Eq. (\ref{H-alpha-appendix}) yields an approximate Hessian, $\tilde{H}_{\bm{\alpha}}$, which corresponds to the diagonal $K \times K$ matrix whose diagonal elements are given by: 
\[ \frac{1}{N} \sum_{i=1}^N (p_{ik} - p_{ik}^2) (\bm{f}_{i}^t\bm{t}_{k})^2, k=1, \dots, K \]
Thus, using $p_{ik} - p_{ik}^2 \leq \frac{1}{4}$, we obtain for $\tau_2 \geq 1$:
\begin{equation}
\lambda_{max}(\tilde{H}_{\bm{\alpha}}) \leq \max_k \frac{\tau_2}{4 N}\sum_{i=1}^N (\bm{f}_{i}^t\bm{t}_{k})^2 
\end{equation}
This gives the approximate Lipschitz constant in Eq. (\ref{Lipschitz-constant-alpha}). 

Finally, it is well-known that, for two blocks of variables with Lipschitz constants $\gamma_{\mathbf{w}}$ and $\gamma_{\bm{\alpha}}$, the following is a Global Lipschitz constant (See \cite{Beck2013}):  
\begin{equation}
\label{Global-Lipschitz-constant-appendix}
\gamma = 2 \max (\gamma_{\mathbf{w}}, \gamma_{\bm{\alpha}})
\end{equation}
Therefore, for $\tau \geq 2$, the expression in Eq. (\ref{Global-Lipschitz-constant}) provides an approximate global Lipschitz constant.  

\section{Proof of Proposition \ref{Approx-minimum-CE-v}} 

Consider the following expressions of $g_1$ and $g_2$, for some $\lambda > 0$:
\begin{align}
\label{g1-and-g2-expressions}
         g_{1} &= -\frac{1}{N}\sum_{i=1}^N\sum_{k=1}^{K}y_{ik}\bm{f}_{i}^{t}(\bm{w}_{k} + \alpha_k \bm{t}_{k})\nonumber +\frac{\lambda}{2} \sum_{k=1}^{K} \|\bm{w}_{k} \|^2\\
     g_{2} &= \frac{1}{N}\sum_{i=1}^{N}\ln \left ( \sum_{j=1}^{K}\exp\bm{f}_{i}^{t}(\bm{w}_{j} + \alpha_j \bm{t}_{j}) \right) -\frac{\lambda}{2} \sum_{k=1}^{K} \|\bm{w}_{k} \|^2
\end{align}
It is easy to verify that $L = g_1 + g_2$. Let us now write the gradients of $g_1$ and $g_2$ w.r.t $\bm{w}_{k}$, $k \in [1, \dots, K]$:
\begin{align}
\label{g1-and-g2-gradient-expressions}
\frac{\partial g_1}{\partial \bm{w}_{k}} &= -\frac{1}{N}\sum_{i=1}^N y_{ik}\bm{f}_{i} + \lambda \bm{w}_{k}  \nonumber \\
\frac{\partial g_2}{\partial \bm{w}_{k}} &= \frac{1}{N}\sum_{i=1}^N p_{ik}\bm{f}_{i} - \lambda \bm{w}_{k} 
\end{align}
Independently of the value of $\lambda > 0$, $g_1$ is convex w.r.t each $\bm{w}_{k}$, $k \in [1, \dots, K]$, as it is the sum of linear and convex functions. 
By setting the gradient of $g_1$ w.r.t $\bm{w}_{k}$ to $0$, we obtain the minimum given in Eq. (\ref{hard-mean-1}). As for $g_{2}$, we can ensure it is convex w.r.t each 
$\mathbf{w}_{k}$ by setting $\lambda$ as a function of positive semi-definite (PSD) matrices $\bm{A}_{k}$, as given in Prop. \ref{Approx-minimum-CE-v}.   
This ensures that the Hessian of $g_2$ is PSD, as $\bm{A}_{k}$ is the Hessian of the first term in $g_2$. Setting the gradient of $g_2$ w.r.t $\bm{w}_{k}$ to $0$ yields the minimum in Eq. (\ref{soft-mean-1}).

\section{More details on the initialization}

The technical observations made in the paper in Eq. (\ref{hard-mean-1}) and Eq. (\ref{hard-mean-alpha}) suggest expressions for initializing variables 
$\alpha_{k}$ and $\bm{w}_{k}$:
\begin{align}
\alpha^0_{k} & = \frac{1}{\beta N} \sum_{i=1}^{N} y_{ik}\bm{f}_{i}^t\bm{t}_{k} = \frac{1}{\beta N} \tilde{\alpha}_{k} \nonumber \\ 
\bm{w}^0_{k} &= \frac{1}{\lambda N}\sum_{i=1}^N y_{ik}\bm{f}_{i} = \frac{1}{\lambda N} \tilde{\bm{w}}_{k} 
\end{align}
There are two non-negative multiplicative factors, $\lambda$ and $\beta$, which appear in these expressions. Here, we 
provide more details on how to set $\lambda$ and $\beta$ systematically (without hyper-parameter search).   

Consider the logit scores for each sample $j$, according to our image-text linear probe model:
\begin{equation}
\label{logit-alpha-beta}
l_{j,k} = \frac{1}{\lambda N} \bm{f}_{j}^t \tilde{\bm{w}}_{k} +  \frac{1}{\beta N} \tilde{\alpha}_{k} \bm{f}_{j}^t \bm{t}_{k} 
\propto  \bm{f}_{j}^t \tilde{\bm{w}}_{k} +   \frac{\lambda}{\beta} \tilde{\alpha}_{k} \bm{f}_{j}^t \bm{t}_{k}  
\end{equation}
Notice that multiplying $\lambda$ and $\beta$ by the same non-negative multiplicative factor does not change the order of the class scores, i.e., we only need to 
set a value for the ratio $\frac{\lambda}{\beta}$, which balances in Eq. (\ref{logit-alpha-beta}) the contribution of the text-embedding term vs. the contribution of the visual-prototype term. We set $\frac{\lambda}{\beta}$ systematically (without additional hyper-parameters) as a monotonically decreasing function of the number of support samples ($S = \frac{N}{K}$):
\begin{equation}
\label{choice-of-alpha-beta-ratio}
\frac{\lambda}{\beta} = \frac{250}{S}
\end{equation}
The choice in Eq. (\ref{choice-of-alpha-beta-ratio}) is motivated by two intuitive reasons. First, the text-embedding scores are at least one-order of magnitude smaller than the visual-embedding scores. Second, a monotonically decreasing function of $S$ makes sense: the smaller the number of visual support samples, the bigger the weight given to the text-knowledge. Finally, by setting $\frac{\lambda}{\beta}$ according to Eq. (\ref{choice-of-alpha-beta-ratio}) and $\lambda = \frac{1}{N}$, we initialize the variables as follows:  
\begin{align}
\label{initialization-paper}
\alpha^0_{k} & = \frac{250}{S} \sum_{i=1}^{N} y_{ik}\bm{f}_{i}^t\bm{t}_{k} \nonumber \\ 
\bm{w}^0_{k} &= \sum_{i=1}^N y_{ik}\bm{f}_{i}
\end{align}

\section{More details on the training-free version of our method}

Here we gave more details on the training-free version of our method LP++. To be specific, for a given test sample, we can compute the class scores (logits) as follows:
\[l_{\textrm{test},k}=\bm{f}^{t}_{\textrm{test}}(\bm{w}^{0}_{k}+\alpha^{0}_{k}\bm{t}_{k}) \quad \forall k\]
where $\bm{f}_\textrm{test}$ is the vision embedding of the test sample and the initial, hyper-parameter free variables $\alpha^0_{k}$ and $\bm{w}^0_{k}$ are 
set according to Eq. (\ref{initialization-paper}). Then, the class of test sample $\bm{f}^{t}_{\textrm{test}}$ is predicted as: 
\[\hat{k} = \argmax_{k} l_{\textrm{test},k} \]


\section{Detailed explanation of the flaw in the experimental evaluation of TIP-adapter \cite{zhang2022tip}}

By examining the official GitHub repository\footnote{\url{https://github.com/gaopengcuhk/Tip-Adapter/issues/13}}, we found that the original implementation of TIP-adapter \cite{zhang2022tip} uses large test-data performance feedback for choosing the hyper-parameters and best model. Therefore, we re-evaluated TIP-adapter using the same small validation sets as those used for all the competing methods, for fairness. Indeed, the paper \cite{zhang2022tip} does not provide details as to the hyper-parameter selection and use of validation data. According to the GitHub repository, the authors mentioned that key hyper-parameters ($\alpha$ and $\beta$ in \cite{zhang2022tip}), which control the trade-off between the text and vision supervision, are both set to $1$ as tuning baseline (see Issue $\#13$ in the GitHub). They also mentioned that $\alpha$ has to be increased when the domain gap between the CLIP pre-trained model and the downstream task is large. Starting from some initial values of the hyper-parameters, with $\beta$ set to 1 and $\alpha \in [1, 10]$ but predetermined specifically for each dataset, the authors deployed an additional search function after the adaptation procedure. {\bf This function finds the best combination of $\alpha$ and $\beta$ on either the test set for ImageNet or the entire validation set for the other 10 datasets, which invalidates the few-shot assumption}. Furthermore, the grid search intervals for $\alpha$ and $\beta$ vary significantly among the different datasets (and somtimes do not even overlap), e.g., $\alpha \in [1.17, 7]$ for ImageNet whereas $\alpha \in [10, 50]$ for Flowers102. 

Through our experiments, we found that the initial value of $\alpha$ plays a substantial role in the final performance of the Tip-Adapter-F model. Hence, for a fair comparison, we re-evaluated Tip-Adapter-F* (Table \ref{table:resultOther_AllDatasets}) by (i) finding the initial value of $\alpha_{\textrm{init}} \in [1, 10]$ on the  small validation set deployed for all methods, and (ii) setting $\beta_{\textrm{init}}=1$. Then, we maintain the implementation of the original publication, and perform the final grid search after adaptation. Instead of employing a data-set specific range for these two hyper-parameters, we use $[\alpha_{\textrm{init}}, 50 ]$ for $\alpha$ and $[\beta_{\textrm{init}}, 50 ]$ for $\beta$ during grid search.
Although these grid-search steps add significantly to the overall computational overhead of TIP-adapter, they are needed to achieve performances close to those reported in the paper \cite{zhang2022tip} (refer to ~\cref{fig:avg-std} for a comparison between Tip-Adapter-F and Tip-Adapter-F*).

As for the training-free version of Tip-adapter \cite{zhang2022tip}, which we evaluated in Table \ref{table:classifier_init}, we fixed these hyper-parameters, i.e., $\alpha=\beta=1$, for a fair comparison with our training-free LP++. Indeed, one could argue that searching for these hyper-parameters on large validation sets would invalidate the claim that the method is `'training-free''.  

In addition to the mentioned flaws, the original implementation of \cite{zhang2022tip} finds the best model using the performance on the test set. In our re-implementation, we addressed this issue by applying early stopping on the small validation set.

\section{Ablation on the block-cycling strategies}

 Table~\ref{table:BMM} reports the performances for different block-cycling strategies in our Block Majorize-Minimize (BMM) procedure. For example, BMM (iter$_{\mathbf{w}}$=10, iter$_{\bm{\alpha}}$=1) corresponds to the LP++ version in Alg. \ref{Algorithm-MM-w-alpha}, whose performance is reported in the paper. GD corresponds to the 
 single-block MM version of LP++, and BCGD (Block Coordinate Gradient Descent) corresponds the block-wise version of LP++ with both iter$_{\mathbf{w}}$ and iter$_{\bm{\alpha}}$ equal to 1.
\begin{table}
\footnotesize
\setlength{\tabcolsep}{4pt}
\renewcommand{\arraystretch}{1.2}
\centering
\begin{tabular}{lccccc}
\hline
\textbf{Number of shots ($S$)} & \textbf{1} & \textbf{2} & \textbf{4} & \textbf{8} & \textbf{16}\\
\hline
\hline
BMM (iter$_{\mathbf{w}}$=10, iter$_{\bm{\alpha}}$=1) & {\bf 63.43} & {\bf 66.20} & {\bf 69.16} & {\bf 72.04} & {\bf 74.42} \\
BMM (iter$_{\mathbf{w}}$=10, iter$_{\bm{\alpha}}$=10) & 62.85 & 66.01 & 69.12 & 71.98 & 74.33 \\
BCGD & 62.92 & 66.04 & 69.03  & 71.94  & 74.24 \\
GD (our Lipschitz cst) & 63.04 & 66.14 & 69.15 & 71.99 &  74.28 \\
\bottomrule
\end{tabular}
\vspace{0.1cm}
\caption{Comparison of different block-cycling strategies using our initialization. BCGD corresponds to BMM with both iter$_{\mathbf{w}}$ and iter$_{\bm{\alpha}}$ equal to 1. GD (our Lipschitz cst) corresponds to the single-block version of LP++. Results are averaged over 11 datasets. 
}
\label{table:BMM}
\end{table}

\section{Method ranking}
\begin{figure}
\centering
\includegraphics[width=0.8\linewidth]{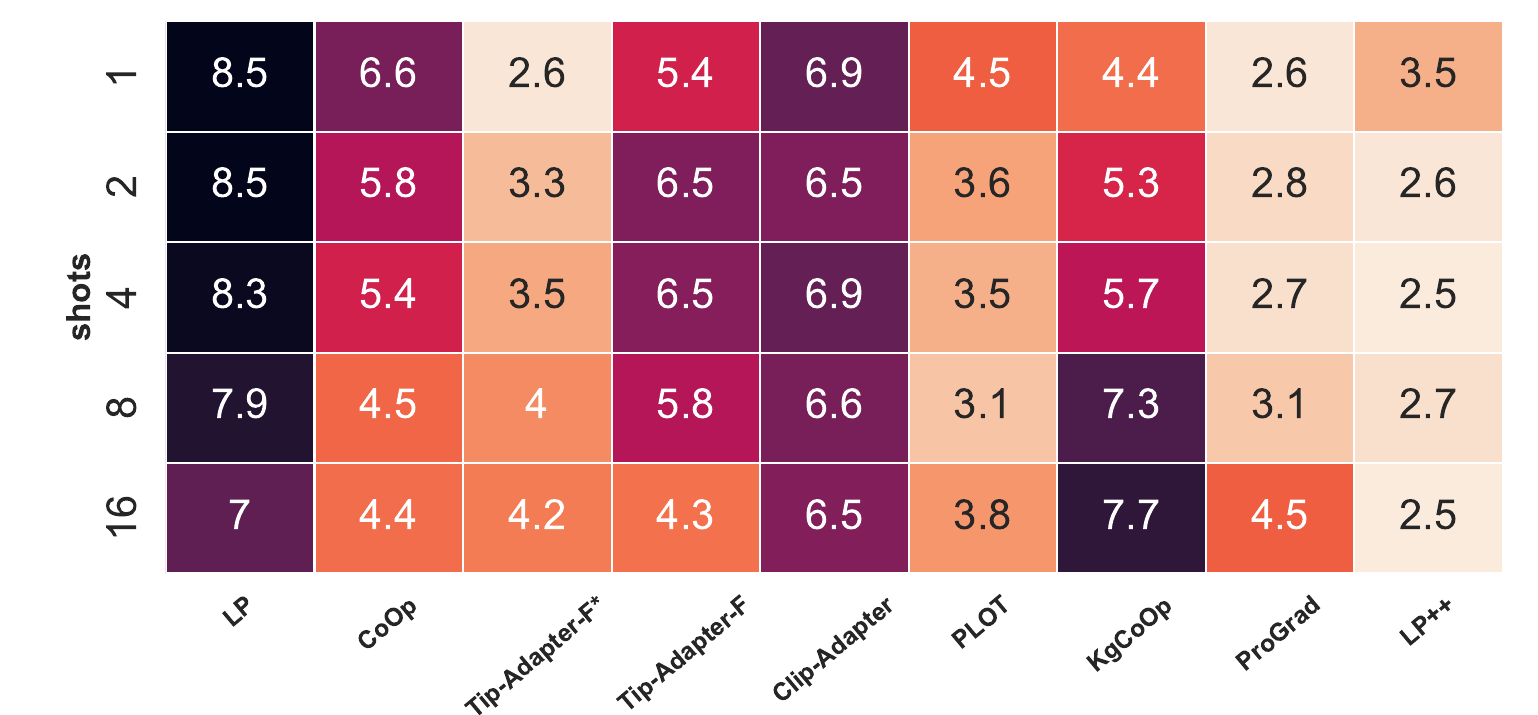}
\caption{The ranking of different methods using Autorank \cite{Herbold2020}. The results are averaged over 11 datasets.}
\label{fig:autorank}
\end{figure}

 In Fig.~\ref{fig:autorank}, we show the ranking of different methods using Autorank\cite{Herbold2020} (on the 11 datasets); lower value indicates better performance.

\section{Implementation of the L-BFGS optimizer}
When assessing the standard LP as in Table~\ref{table:resultOther_AllDatasets_details}, we follow the instructions provided in \cite{Zhou2022coop}, where the L-BFGS optimizer is implemented by NumPy, and used to optimize the standard CE loss jointly with L2 regularizers. As for the ablation study in Table~\ref{table:ablation-loss-and-optimizers}, all the optimizers are implemented by Pytorch, including L-BFGS\footnote{  \url{https://github.com/hjmshi/PyTorch-LBFGS}}. In this case, we did not use regularizers, for fair comparisons.

\section{Details of prompt design}
We follow the same strategy as in \cite{zhang2022tip} for the text prompts. To be specific, we use the average of 7 templates per class for ImageNet, while for the other datasets, we use a single template. The latter might be different according to the type of the images those datasets include. For example, for Caltech101, we simply use ``a photo of a [class$_{k}$]'' for each class, while for Food101, we use ``a photo of [class$_{k}$], a type of food'' for each class.

\begin{figure*}
     \centering

     \includegraphics[width=0.9\linewidth]{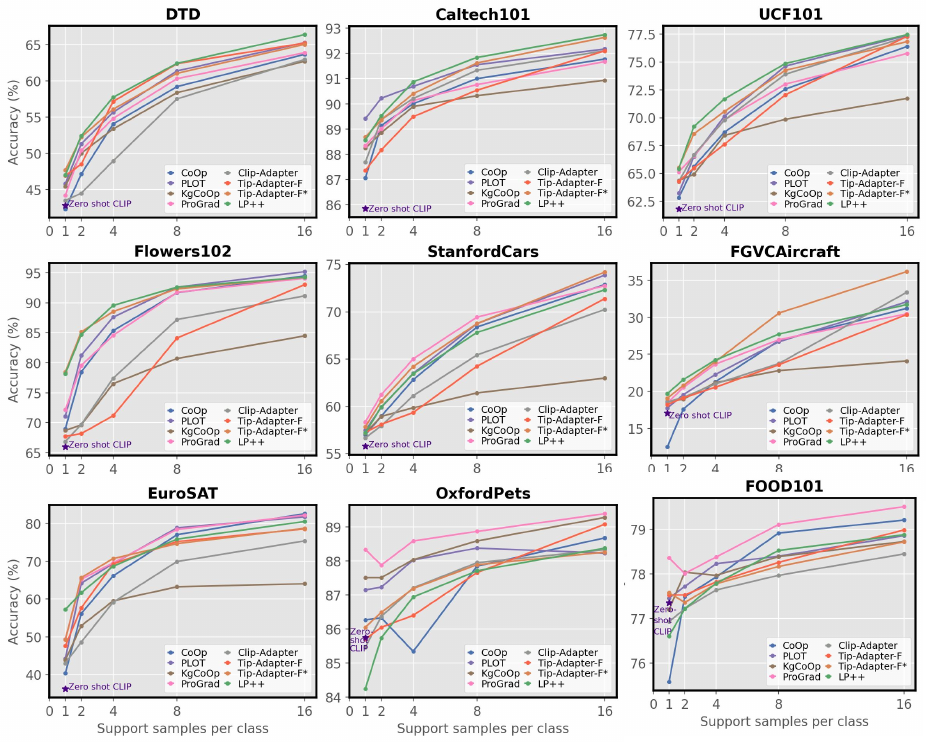}

    \caption{Comparison of LP++ with different adaptation methods on 9 benchmarks, which are averaged over 10 tasks.}
    \label{fig:main2-sup}
\end{figure*}

\section{Proof Lemma 2.1.}

Assume a function $L(\mathbf{v})$ is $\gamma$-smooth, i.e., its gradient is Lipschitz (with a Lipschitz constant $\gamma>0$): $\nabla^2 L(\mathbf{v}) \preceq \gamma \mathbf{I}$, with $\mathbf{I}$ the identity matrix.  
From this condition on the Hessian of $L$, it is easy to see that the following function is convex:
$G(\mathbf{v}) = \frac{\gamma}{2} \mathbf{v}^t \mathbf{v} - L(\mathbf{v})$.
Using the first-order condition for convexity of $G$ gives: 
\begin{equation}
\label{first-order-convexity}
G(\mathbf{v}) \geq G(\mathbf{v}^j) + \nabla G(\mathbf{v}^j)^t (\mathbf{v} - \mathbf{v}^j)
\end{equation}
Then using the expression of $G$ and its gradient in Eq. (\ref{first-order-convexity}) yield the following quadratic upper bound on $L$, which 
corresponds to majorizing function $M(\mathbf{v},\mathbf{v}^j)$ in the paper:
\begin{equation}
\label{quadratic-bound-appendix}
L(\mathbf{v}) \leq  L(\mathbf{v}^j) + \nabla L(\mathbf{v}^j)^t(\mathbf{v}-\mathbf{v}^j) + \frac{\gamma}{2} \|\mathbf{v}-\mathbf{v}^j \|^2
\end{equation}
By setting the derivative of this convex upper bound to zero, we get the following solution, which corresponds to 
a specific gradient-descent step: 
\begin{equation}
\label{gradient-step-specific}
\mathbf{v}^{j+1} = \mathbf{v}^{j} - \frac{1}{\gamma} \nabla L(\mathbf{v}^{j}) = \arg \min_{\mathbf{v}} {M(\mathbf{v}, \mathbf{v}^j)}
\end{equation}
Applying Eq. (\ref{quadratic-bound-appendix}) to $\mathbf{v} = \mathbf{v}^{j+1}$, and plugging the gradient step in Eq. (\ref{gradient-step-specific}) 
into Eq. (\ref{quadratic-bound-appendix}), one gets the following decrease of objective $L$, by at least $\frac{1}{2\gamma} \|\nabla L(\mathbf{v}^j)\|^2$:
\begin{equation}
\label{descent-Lemma-Appendix}
L(\mathbf{v}^{j+1}) \leq L(\mathbf{v}^{j}) - \frac{1}{2\gamma} \|\nabla L(\mathbf{v}^{j})\|^2 
\end{equation}

\section{Proof Theorem 2.2.}

The standard result in Theorem 2.2 provides, for a convex, $\gamma$-smooth function, the convergence rate of gradient steps with step
sizes $\frac{1}{\gamma}$. As $L$ is convex, we have: 
\begin{equation}
\label{first-order-bound-convex}
 L(\mathbf{v}^{j}) \leq L(\mathbf{v}^{*}) + \nabla L(\mathbf{v}^j)^t(\mathbf{v}^{j}-\mathbf{v}^*) 
\end{equation}
Plugging Eq. (\ref{first-order-bound-convex}) in Eq. (\ref{descent-Lemma-Appendix}) yields:
\begin{align}
\label{first-bound-optimal-value}
& L(\mathbf{v}^{j+1}) - L(\mathbf{v}^{*}) \\
& \leq   \nabla L(\mathbf{v}^j)^t(\mathbf{v}^{j}-\mathbf{v}^*) \nonumber  - \frac{1}{2\gamma} \|\nabla L(\mathbf{v}^{j})\|^2 \nonumber \\
& = \frac{\gamma}{2}  ( \|\mathbf{v}^j - \mathbf{v}^*\|^2 - \|\mathbf{v}^j - \frac{1}{\gamma} \nabla L(\mathbf{v}^{j}) \nonumber -  \mathbf{v}^* \|^2 )  \nonumber \\
&=\frac{\gamma}{2}  ( \|\mathbf{v}^j - \mathbf{v}^*\|^2 -  \|\mathbf{v}^{j+1} - \mathbf{v}^*\|^2 )
\end{align}
Summing both parts of the inequality in Eq. (\ref{first-bound-optimal-value}), we get:
\begin{align}
\label{second-bound-optimal-value}
&\sum_{j=0}^{J-1} \left(L(\mathbf{v}^{j+1}) - L(\mathbf{v}^{*})\right)\\
&\leq \frac{\gamma}{2} \sum_{j=0}^{J-1}   ( \|\mathbf{v}^j - \mathbf{v}^*\|^2 \nonumber 
-  \|\mathbf{v}^{j+1} - \mathbf{v}^*\|^2 ) \nonumber \\ & = \frac{\gamma}{2} ( \|\mathbf{v}^0 - \mathbf{v}^*\|^2 - \|\mathbf{v}^J - \mathbf{v}^*\|^2 )
\nonumber \\ & \leq  \frac{\gamma}{2} ( \|\mathbf{v}^0 - \mathbf{v}^*\|^2 )
\end{align}
Finally, using the fact that $L$ is decreasing at each iteration, along with inequality Eq. (\ref{second-bound-optimal-value}), we have: 
\begin{align}
L(\mathbf{v}^{J}) - L(\mathbf{v}^{*}) &\leq \frac{1}{J} \sum_{j=0}^{J-1} \left(L(\mathbf{v}^{j+1}) - L(\mathbf{v}^{*})\right) \nonumber\\
&\leq \frac{\gamma}{2J} ( \|\mathbf{v}^0 - \mathbf{v}^*\|^2 ) 
\end{align}




\section{Additional numerical results on 11 benchmarks}
\label{sec:additional_results}
~\cref{table:resultOther_AllDatasets_details} shows the average accuracy and standard deviation over 10 tasks for LP++ and the baseline methods on 11 benchmarks. The per-dataset performance across all the methods and for different number of shots 
is depicted in ~\cref{fig:main2-sup}, which complements \cref{fig:clip_mag} in the main paper. In these plots, one may observe that LP++ (\textit{green curve}) yields the best overall performance in 6 out of 11 data sets. Furthermore, LP++ outperforms all adapter-based methods in nearly all the remaining datasets and scenarios (i.e., when varying the number of shots). Only in a few cases, e.g., FGVCAirCraft and OxfordPets, Tip-Adapter-F* yields a performance better than LP++. Indeed, the largest improvement gains are observed in FGVCAirCraft, where the accuracy obtained by all methods is considerably small. We believe that, in this particular scenario, the more intensive grid search carried out by Tip-Adapter-F* has a greater impact on its results. Considering 
prompt learning strategies, ProGrad is the top-performing approach, which obtains the best overall results in several datasets. Nevertheless, in addition to the limitations already highlighted for these methods (i.e., time complexity and non-suitability for black-box adaptation), the large standard deviations observed across runs (see also \cref{fig:avg-std}), make these approaches highly unreliable, specially in the low-labeled data regime. As discussed, this could be due to the fact that these approaches learn prompts, through the text encoder, which are 
``too specialized'' for a given image support set.

\begin{table*}[ht] 
\footnotesize
\tiny
\centering
\renewcommand{\arraystretch}{1.2}
\begin{tabular}{l|lccccc}
\hline
 Dataset & \textbf{Number of shots ($S$)} & \textbf{1} & \textbf{2} & \textbf{4} & \textbf{8} & \textbf{16}\\
\hline
\hline
\multirow{10}{*}{\textit{ImageNet}} & Zero-shot CLIP$_\text{ ICML'21}$\cite{radford2021clip} & \multicolumn{5}{c}{60.35}\\
\cline{2-7}
& CoOp$_\text{ IJCV'22}$\cite{Zhou2022coop} & $61.19 \pm 0.17$ & $61.58 \pm 0.40$  &  $62.22 \pm 0.22$  & $62.87 \pm 0.21$  &  $63.70 \pm 0.13$ \\
& PLOT$_\text{ ICLR'23}$\cite{chen2023plot} & $60.46 \pm 0.34$ &  $60.73 \pm 0.60$ &  $61.79 \pm 0.39$  &  $62.48 \pm 0.32$ & $63.08 \pm 0.26$  \\
& KgCoOp$_\text{ CVPR'23}$\cite{Yao2023kgcoop} & $60.90 \pm 0.16$ & $61.44 \pm 0.15$ & $62.00 \pm 0.11$ &$62.20 \pm 0.15$ & $62.43 \pm 0.11$ \\
& ProGrad$_\text{ ICCV'23}$\cite{Zhu2023prograd} & $\bm{61.58} \pm 0.27$ & $\bm{62.14} \pm 0.13$ & $\bm{62.59} \pm 0.09$ & $63.04 \pm 0.11$ & $63.54 \pm 0.18$\\
\cline{2-7}
& CLIP-Adapter$_\text{ IJCV'23}$\cite{Gao2023clipadapter} & $59.82 \pm 0.11 $ & $59.94 \pm 0.05$ & $59.97 \pm 0.04$ & $59.98 \pm 0.09$ & $ 61.31 \pm 0.39$ \\
& Tip-Adapter-F$_\text{ ECCV'22}$\cite{zhang2022tip} & $60.59 \pm0.14$ & $61.42 \pm0.05$ & $62.12 \pm0.06$  & $63.41\pm 0.07$  & $\bm{65.06}\pm 0.04$ \\
& Tip-Adapter-F*$_\text{ ECCV'22}$\cite{zhang2022tip} & $60.98 \pm0.15$ & $61.23 \pm0.11$ & $61.72 \pm0.25$ & $62.84\pm 0.10$ & $64.03\pm 0.12$ \\
\cline{2-7}
& Standard LP$_\text{ ICML'21}$\cite{radford2021clip} & $22.21 \pm 0.31$  &  $31.96 \pm0.25$  &  $41.48 \pm0.25$  &  $49.49 \pm0.16$  & $56.04 \pm0.13$  \\
& \cellcolor{blue!15}{LP++} & \cellcolor{blue!15}{$61.18 \pm 0.08$} &  \cellcolor{blue!15}{$61.56 \pm 0.14$} & \cellcolor{blue!15}{$62.55 \pm 0.12$} & \cellcolor{blue!15}{$\bm{63.76} \pm 0.07$} & \cellcolor{blue!15}{$64.73 \pm  0.05$}\\
\midrule

\multirow{10}{*}{\textit{SUN397}} & Zero-shot CLIP$_\text{ ICML'21}$\cite{radford2021clip} & \multicolumn{5}{c}{58.85}\\
\cline{2-7}
& CoOp$_\text{ IJCV'22}$\cite{Zhou2022coop} &$61.79 \pm 0.45$ &  $63.32 \pm 0.47$ & $65.79 \pm 0.44$ & $67.89 \pm 0.38$  &  $70.15 \pm 0.32$ \\
& PLOT$_\text{ ICLR'23}$\cite{chen2023plot} & $62.53 \pm 0.30$ & $63.87 \pm 0.26$ & $65.85 \pm 0.48$ & $67.83 \pm 0.36$ & $69.90 \pm 0.31$  \\
& KgCoOp$_\text{ CVPR'23}$\cite{Yao2023kgcoop} & $\bm{62.91} \pm 0.59$  &  $64.38 \pm 0.30$ & $66.06 \pm 0.37$ & $66.66 \pm 1.10$ & $67.68 \pm 0.78$ \\
& ProGrad$_\text{ ICCV'23}$\cite{Zhu2023prograd} & $62.79 \pm 0.50$ & $64.12 \pm 0.55$ & $66.32 \pm 0.59$ & $68.33 \pm 0.28$ & $70.18 \pm 0.27$\\
\cline{2-7}
& CLIP-Adapter$_\text{ IJCV'23}$\cite{Gao2023clipadapter} & $60.78 \pm 0.16 $ & $61.79 \pm 0.23$ & $63.84 \pm 0.35$ & $66.26 \pm 0.14$ & $ 67.66 \pm 1.05$ \\
& Tip-Adapter-F$_\text{ ECCV'22}$\cite{zhang2022tip} & $61.02 \pm 0.36$  & $62.15 \pm 0.28$  & $63.86 \pm 0.19$ & $67.25 \pm 0.16$ & $70.94 \pm 0.13$ \\
& Tip-Adapter-F*$_\text{ ECCV'22}$\cite{zhang2022tip} & $62.58 \pm 0.22$  & $63.79 \pm 0.13$  & $65.49 \pm 0.35$ & $67.43 \pm 0.11$ & $69.25 \pm 0.16$ \\
\cline{2-7}
& Standard LP$_\text{ ICML'21}$\cite{radford2021clip} & $32.56 \pm 0.40$  & $43.77 \pm 0.41$   &  $54.49 \pm 0.39$  &  $61.83 \pm 0.30$  &  $67.03 \pm 0.16$  \\
& \cellcolor{blue!15}{LP++} & \cellcolor{blue!15}{$62.47 \pm 0.27$} & \cellcolor{blue!15}{$\bm{64.65} \pm 0.25$} & \cellcolor{blue!15}{$\bm{67.28} \pm 0.27$} & \cellcolor{blue!15}{$\bm{69.34} \pm 0.14$} & \cellcolor{blue!15}{$\bm{71.23} \pm 0.07$}\\
\midrule

\multirow{10}{*}{\textit{DTD}} & Zero-shot CLIP$_\text{ ICML'21}$\cite{radford2021clip} & \multicolumn{5}{c}{42.69}\\
\cline{2-7}
& CoOp$_\text{ IJCV'22}$\cite{Zhou2022coop} & $42.31 \pm 1.86$ &  $47.13 \pm 1.93$ &  $54.06 \pm 1.49$  & $59.21 \pm 0.91$  &  $63.67 \pm 0.83$ \\
& PLOT$_\text{ ICLR'23}$\cite{chen2023plot} & $45.82 \pm 1.72$ & $51.32 \pm 1.61$  & $55.67 \pm 1.14$   &  $61.38 \pm 1.04$ &  $65.29 \pm 1.05$  \\
& KgCoOp$_\text{ CVPR'23}$\cite{Yao2023kgcoop} & $45.46 \pm 2.83$ & $50.01 \pm 2.71$ & $53.37 \pm 0.71$ &  $58.38 \pm 1.34$  & $62.71 \pm 0.92$ \\
& ProGrad$_\text{ ICCV'23}$\cite{Zhu2023prograd} & $44.19 \pm 2.38$  & $ 50.41 \pm 1.74$  & $54.82 \pm 1.28$ & $60.31 \pm 0.99$ & $63.89 \pm 0.88$\\
\cline{2-7}
& CLIP-Adapter$_\text{ IJCV'23}$\cite{Gao2023clipadapter} & $43.49 \pm 0.68 $ & $44.49 \pm 1.07$ & $48.95 \pm 0.85$ & $57.52 \pm 0.67$ & $ 62.97 \pm 0.60$ \\
& Tip-Adapter-F$_\text{ ECCV'22}$\cite{zhang2022tip} & $46.92 \pm 1.01$ & $48.50 \pm 1.08$ & $57.16 \pm 0.53$ & $62.38 \pm 0.47$ & $65.23 \pm 0.82$ \\
& Tip-Adapter-F*$_\text{ ECCV'22}$\cite{zhang2022tip} & $\bm{47.68} \pm 1.43$ & $52.24 \pm 0.74$ & $56.09 \pm 0.99$ & $61.05 \pm 0.71$ & $65.04 \pm 0.21$ \\
\cline{2-7}
& Standard LP$_\text{ ICML'21}$\cite{radford2021clip} & $29.63 \pm 1.53$ & $41.19 \pm 1.45$   & $51.72 \pm 0.57$   &  $58.78 \pm 0.52$  & $64.56 \pm 0.69$  \\
& \cellcolor{blue!15}{LP++} & \cellcolor{blue!15}{$46.97 \pm 1.37$} & \cellcolor{blue!15}{$\bm{52.44} \pm 0.99$} & \cellcolor{blue!15}{$\bm{57.75} \pm 0.82$} & \cellcolor{blue!15}{$\bm{62.42} \pm 0.53$} & \cellcolor{blue!15}{$\bm{66.40} \pm 0.50$}\\
\midrule

\multirow{10}{*}{\textit{Caltech101}} & Zero-shot CLIP$_\text{ ICML'21}$\cite{radford2021clip} & \multicolumn{5}{c}{85.84}\\
\cline{2-7}
& CoOp$_\text{ IJCV'22}$\cite{Zhou2022coop} & $87.06 \pm 1.24$ & $89.14 \pm 0.87$  &  $90.00 \pm 0.63$  &  $91.00 \pm 0.66$ &  $91.77 \pm 0.29$ \\
& PLOT$_\text{ ICLR'23}$\cite{chen2023plot} & $\bm{89.41} \pm 0.41$ & $\bm{90.22} \pm 0.25$  &  $90.69 \pm 0.37$  &  $91.55 \pm 0.38$ &  $92.17 \pm 0.30$  \\
& KgCoOp$_\text{ CVPR'23}$\cite{Yao2023kgcoop} & $88.24 \pm 0.49$ & $88.85 \pm 0.43$  &  $89.89 \pm 0.31$ & $90.32 \pm 0.43$ & $90.93 \pm 0.26$ \\
& ProGrad$_\text{ ICCV'23}$\cite{Zhu2023prograd} & $88.34 \pm 1.64$  & $89.01 \pm 0.61$  & $90.13 \pm 0.45$  & $90.76 \pm 0.32$ & $91.67 \pm 0.39$\\
\cline{2-7}
& CLIP-Adapter$_\text{ IJCV'23}$\cite{Gao2023clipadapter} & $87.69 \pm 0.41 $ & $89.37 \pm 0.29$ & $90.21 \pm 0.32$ & $91.33 \pm 0.15$ & $ 92.10 \pm 0.20$ \\
& Tip-Adapter-F$_\text{ ECCV'22}$\cite{zhang2022tip} & $87.35 \pm 0.64$ & $88.17 \pm 0.29$ & $89.49 \pm 0.25$ & $90.54 \pm 0.34$ & $92.10 \pm 0.25$ \\
& Tip-Adapter-F*$_\text{ ECCV'22}$\cite{zhang2022tip} & $88.68 \pm 0.44$ & $89.36 \pm 0.59$ & $90.40 \pm 0.26$ & $91.62 \pm 0.23$ & $92.63 \pm 0.21$ \\
\cline{2-7}
& Standard LP$_\text{ ICML'21}$\cite{radford2021clip} & $68.88 \pm 1.68$  &  $78.41 \pm 0.54$  &  $84.91 \pm 0.45$  &  $88.70 \pm 0.40$  &  $91.14 \pm 0.19$  \\
& \cellcolor{blue!15}{LP++} & \cellcolor{blue!15}{$88.56 \pm 0.43$} & \cellcolor{blue!15}{$89.53 \pm 0.35$} & \cellcolor{blue!15}{$\bm{90.87} \pm 0.19$} & \cellcolor{blue!15}{$\bm{91.84} \pm 0.24$} & \cellcolor{blue!15}{$\bm{92.73} \pm 0.17$}\\
\midrule

\multirow{10}{*}{\textit{UCF101}} & Zero-shot CLIP$_\text{ ICML'21}$\cite{radford2021clip} & \multicolumn{5}{c}{61.80}\\
\cline{2-7}
& CoOp$_\text{ IJCV'22}$\cite{Zhou2022coop} & $62.80 \pm 1.26$ &  $65.62 \pm 1.09$ & $68.69 \pm 0.76$  &  $72.57 \pm 0.80$ & $76.39 \pm 0.54$ \\
& PLOT$_\text{ ICLR'23}$\cite{chen2023plot} & $63.22 \pm 1.05$ & $66.49 \pm 0.92$& $70.12 \pm 0.62$& $74.63 \pm 0.79$ & $77.39 \pm 0.53$  \\
& KgCoOp$_\text{ CVPR'23}$\cite{Yao2023kgcoop} & $64.37 \pm 1.66$  & $64.91 \pm 1.01$ & $68.41 \pm 0.38$  & $69.86 \pm 0.33$ &  $71.73 \pm 0.78$ \\
& ProGrad$_\text{ ICCV'23}$\cite{Zhu2023prograd} & $65.13 \pm 0.87$  & $66.57 \pm 0.62$ & $69.80 \pm 0.62$ & $73.01 \pm 0.52$ & $75.76 \pm 0.47$\\
\cline{2-7}
& CLIP-Adapter$_\text{ IJCV'23}$\cite{Gao2023clipadapter} & $64.25 \pm 0.54 $ & $66.68 \pm 0.31$ & $69.77 \pm 0.40$ & $73.90 \pm 0.50$ & $ 77.26 \pm 0.39$ \\
& Tip-Adapter-F$_\text{ ECCV'22}$\cite{zhang2022tip} & $64.28 \pm 0.54$ & $65.48 \pm 0.43$ & $67.61 \pm 0.28$ & $72.05 \pm 0.53$ & $77.30 \pm 0.21$  \\
& Tip-Adapter-F*$_\text{ ECCV'22}$\cite{zhang2022tip} & $\bm{65.50} \pm 0.59$ & $68.55 \pm 0.45$ & $70.55 \pm 0.58$ & $74.25 \pm 0.48$ & $76.83 \pm 0.24$  \\
\cline{2-7}
& Standard LP$_\text{ ICML'21}$\cite{radford2021clip} & $40.80 \pm 1.05$ &  $51.71 \pm 0.79$  &  $61.64 \pm 0.50$  &  $68.47 \pm 0.44$  &  $73.38 \pm 0.43$  \\
& \cellcolor{blue!15}{LP++} & \cellcolor{blue!15}{$65.41 \pm 0.37$} & \cellcolor{blue!15}{$\bm{69.20} \pm 0.52$} & \cellcolor{blue!15}{$\bm{71.68} \pm 0.41$} & \cellcolor{blue!15}{$\bm{74.86} \pm 0.36$} & \cellcolor{blue!15}{$\bm{77.46} \pm 0.39$}\\
\midrule

\multirow{10}{*}{\textit{Flowers102}} & Zero-shot CLIP$_\text{ ICML'21}$\cite{radford2021clip} & \multicolumn{5}{c}{65.98}\\
\cline{2-7}
& CoOp$_\text{ IJCV'22}$\cite{Zhou2022coop} & $69.00 \pm 2.44$ & $78.47 \pm 1.88$  & $85.34 \pm 1.69$   &  $91.68 \pm 0.82$ & $94.47 \pm 0.36$ \\
& PLOT$_\text{ ICLR'23}$\cite{chen2023plot} & $71.09 \pm 1.44$ &  $81.22 \pm 0.92$ &  $87.61 \pm 0.79$  &  $92.60 \pm 0.55$ &  $\bm{95.18} \pm 0.40$  \\
& KgCoOp$_\text{ CVPR'23}$\cite{Yao2023kgcoop} & $68.73 \pm 1.97$ & $69.63 \pm 1.25$ &  $76.51 \pm 0.51$ & $80.71 \pm 0.63$ & $84.48 \pm 0.70$ \\
& ProGrad$_\text{ ICCV'23}$\cite{Zhu2023prograd} & $72.16 \pm 1.74$ & $79.55 \pm 0.88$ & $84.56 \pm 1.41$ & $91.73 \pm 0.35$ &  $94.10 \pm 0.41$\\
\cline{2-7}
& CLIP-Adapter$_\text{ IJCV'23}$\cite{Gao2023clipadapter} & $66.86 \pm 0.73 $ & $69.71 \pm 0.46$ & $77.42 \pm 0.60$ & $87.20 \pm 0.52$ & $ 91.16 \pm 0.23$ \\
& Tip-Adapter-F$_\text{ ECCV'22}$\cite{zhang2022tip} & $67.73 \pm 0.57$ & $68.18 \pm 0.84$ & $71.17 \pm 0.67$ & $84.11 \pm 0.49$ & $93.02 \pm 0.28$ \\
& Tip-Adapter-F*$_\text{ ECCV'22}$\cite{zhang2022tip} & $\bm{78.46} \pm 1.01$ & $\bm{85.14} \pm 0.81$ & $88.53 \pm 0.54$ & $92.33 \pm 0.32$ & $94.26 \pm 0.38$ \\
\cline{2-7}
& Standard LP$_\text{ ICML'21}$\cite{radford2021clip} & $56.98 \pm 1.56$  &  $73.40 \pm 0.87$  &  $84.38 \pm 0.53$  &  $91.81 \pm 0.34$  & $95.05 \pm 0.29$  \\
& \cellcolor{blue!15}{LP++} & \cellcolor{blue!15}{$78.21 \pm 1.01$} & \cellcolor{blue!15}{$84.69 \pm 0.70$} & \cellcolor{blue!15}{$\bm{89.56} \pm 0.45$} & \cellcolor{blue!15}{$\bm{92.61} \pm 0.32$} & \cellcolor{blue!15}{$94.26 \pm 0.24$}\\
\midrule

\multirow{10}{*}{\textit{StanfordCars}} & Zero-shot CLIP$_\text{ ICML'21}$\cite{radford2021clip} & \multicolumn{5}{c}{55.78}\\
\cline{2-7}
& CoOp$_\text{ IJCV'22}$\cite{Zhou2022coop} & $57.00 \pm 0.93$& $58.96 \pm 0.78$ & $62.81 \pm 0.71$ & $68.40 \pm 0.61$ & $72.87 \pm 0.50$ \\
& PLOT$_\text{ ICLR'23}$\cite{chen2023plot} & $57.47 \pm 0.58$& $59.89 \pm 0.60$& $63.49 \pm 0.80$&  $68.75 \pm 0.46$ &  $73.86 \pm 0.39$  \\
& KgCoOp$_\text{ CVPR'23}$\cite{Yao2023kgcoop} & $57.19 \pm 0.65$  & $58.94 \pm 0.33$ & $59.85 \pm 0.51$ & $61.42 \pm 0.40$ & 
 $62.99 \pm 1.33$ \\
& ProGrad$_\text{ ICCV'23}$\cite{Zhu2023prograd} & $\bm{58.63} \pm 0.39$ & $\bm{61.23} \pm 0.65$ & $\bm{65.02} \pm 0.78$ & $\bm{69.43} \pm 0.40$ & $72.76 \pm 0.45$\\
\cline{2-7}
& CLIP-Adapter$_\text{ IJCV'23}$\cite{Gao2023clipadapter} & $56.67 \pm 0.22 $ & $57.94 \pm 0.27$ & $61.13 \pm 0.30$ & $65.43 \pm 0.10$ & $ 70.24 \pm 0.79$ \\
& Tip-Adapter-F$_\text{ ECCV'22}$\cite{zhang2022tip} & $57.24 \pm 0.23$ & $58.12 \pm 0.50$ & $59.34 \pm 0.20$ & $64.25 \pm 0.19$ & $71.38 \pm 0.23$  \\
& Tip-Adapter-F*$_\text{ ECCV'22}$\cite{zhang2022tip} & $57.85 \pm 0.33$ & $60.55 \pm 0.34$ & $64.22 \pm 0.52$ & $68.75 \pm 0.31$ & $\bm{74.19} \pm 0.30$ \\
\cline{2-7}
& Standard LP$_\text{ ICML'21}$\cite{radford2021clip} & $22.94 \pm 0.61$  &  $35.48 \pm 0.51$  &  $47.49 \pm 0.67$  &  $59.34 \pm 0.30$  & $69.11 \pm 0.18$  \\
& \cellcolor{blue!15}{LP++} & \cellcolor{blue!15}{$57.20 \pm 0.65$} & \cellcolor{blue!15}{$59.95 \pm 0.36$} & \cellcolor{blue!15}{$63.44 \pm 0.34$} & \cellcolor{blue!15}{$67.81 \pm 0.24$} & \cellcolor{blue!15}{$72.33 \pm 0.18$}\\
\midrule
\end{tabular}
    \vspace{0.1cm}
    \caption{\textbf{Comparison to state-of-the-art methods.} Average classification accuracy (\%) and standard deviation over 10 tasks for 11 benchmarks, 
    Best values are highlighted in bold.}
    \label{table:resultOther_AllDatasets_details}
\end{table*}

\begin{table*}[ht] 
\footnotesize
\tiny
\centering
\renewcommand{\arraystretch}{1.2}
\begin{tabular}{l|lccccc}
\hline
 Dataset & \textbf{Number of shots ($S$)} & \textbf{1} & \textbf{2} & \textbf{4} & \textbf{8} & \textbf{16}\\
\hline
\hline
\multirow{10}{*}{\textit{FGVCAircraft}} & Zero-shot CLIP$_\text{ ICML'21}$\cite{radford2021clip} & \multicolumn{5}{c}{17.07}\\
\cline{2-7}
& CoOp$_\text{ IJCV'22}$\cite{Zhou2022coop} & $12.50 \pm 6.16$& $17.59 \pm 3.70$ & $21.27 \pm 2.54$& $26.85 \pm 0.63$& $31.20 \pm 0.40$ \\
& PLOT$_\text{ ICLR'23}$\cite{chen2023plot} & $17.75 \pm 1.36$ & $19.55 \pm 0.99$ & $22.26 \pm 0.89$& $26.70 \pm 0.70$  & $32.09 \pm 0.68$  \\
& KgCoOp$_\text{ CVPR'23}$\cite{Yao2023kgcoop} & $ 18.61 \pm 0.76$  &  $18.93 \pm 1.01$ & $21.16 \pm 0.82$ & $22.80 \pm 0.44$  & $24.10 \pm 0.59$ \\
& ProGrad$_\text{ ICCV'23}$\cite{Zhu2023prograd} & $18.41 \pm 0.98$  & $20.51 \pm 1.11$ & $23.65 \pm 0.50$ & $26.98 \pm 0.50$ & $30.47 \pm 0.76$\\
\cline{2-7}
& CLIP-Adapter$_\text{ IJCV'23}$\cite{Gao2023clipadapter} & $18.56 \pm 0.20 $ & $19.18 \pm 0.28$ & $21.00 \pm 0.21$ & $23.76 \pm 0.33$ & $ 33.37 \pm 0.23$ \\
& Tip-Adapter-F$_\text{ ECCV'22}$\cite{zhang2022tip} & $18.23 \pm 0.19$ & $19.12 \pm 0.20$ & $20.55 \pm 0.20$ & $23.60 \pm 0.29$ & $30.37 \pm 0.25$ \\
& Tip-Adapter-F*$_\text{ ECCV'22}$\cite{zhang2022tip} & $19.08 \pm 0.15$ & $20.79 \pm 0.59$ & $23.99 \pm 0.57$ & $\bm{30.58} \pm 0.29$ & $\bm{36.16} \pm 0.34$ \\
\cline{2-7}
& Standard LP$_\text{ ICML'21}$\cite{radford2021clip} & $12.66 \pm 0.59$ &  $16.92 \pm 0.56$  &  $21.11 \pm 0.83$  &  $26.53 \pm 0.38$  &  $32.42 \pm 0.54$  \\
& \cellcolor{blue!15}{LP++} & \cellcolor{blue!15}{$\bm{19.69} \pm 0.39$} & \cellcolor{blue!15}{$\bm{21.58} \pm 0.46$} & \cellcolor{blue!15}{$\bm{24.22} \pm 0.60$} & \cellcolor{blue!15}{$27.73 \pm 0.48$} & \cellcolor{blue!15}{$31.73 \pm 0.44$}\\
\midrule
\multirow{10}{*}{\textit{EuroSAT}} & Zero-shot CLIP$_\text{ ICML'21}$\cite{radford2021clip} & \multicolumn{5}{c}{36.22}\\
\cline{2-7}
& CoOp$_\text{ IJCV'22}$\cite{Zhou2022coop} & $40.36 \pm 7.19$& $56.15 \pm 5.82$& $66.13 \pm 3.62$&  $77.02 \pm 1.78$ & $\bm{82.59} \pm 1.00$ \\
& PLOT$_\text{ ICLR'23}$\cite{chen2023plot} & $44.22 \pm 9.14$& $64.19 \pm 6.24$ & $69.37 \pm 3.26$&  $\bm{78.84} \pm 1.33$ & $81.76 \pm 1.43$  \\
& KgCoOp$_\text{ CVPR'23}$\cite{Yao2023kgcoop} & $43.86 \pm 9.17$  &  $ 52.92 \pm 5.92$  &  $ 59.51 \pm 3.46$  &  $63.23 \pm 3.03$   &  $64.04 \pm 1.40$ \\
& ProGrad$_\text{ ICCV'23}$\cite{Zhu2023prograd} & $49.37 \pm 5.03$ & $65.22 \pm 4.01$ & $69.57 \pm 2.88$  & $78.44 \pm 1.69$ & $82.17 \pm 0.98$\\
\cline{2-7}
& CLIP-Adapter$_\text{ IJCV'23}$\cite{Gao2023clipadapter} & $43.00 \pm 2.27 $ & $48.60 \pm 2.76$ & $59.15 \pm 2.26$ & $69.92 \pm 1.49$ & $ 75.38 \pm 0.78$ \\
& Tip-Adapter-F$_\text{ ECCV'22}$\cite{zhang2022tip} & $47.63 \pm 2.64$ & $57.62 \pm 1.86$ & $69.30 \pm 2.41$ & $75.22 \pm 1.32$ & $78.59 \pm 1.48$ \\
& Tip-Adapter-F*$_\text{ ECCV'22}$\cite{zhang2022tip} & $49.27 \pm 2.88$ & $\bm{65.66} \pm 1.39$ & $\bm{70.72} \pm 2.73$ & $74.66 \pm 3.15$ & $78.73 \pm 0.81$ \\
\cline{2-7}
& Standard LP$_\text{ ICML'21}$\cite{radford2021clip} & $48.29 \pm 2.95$  & $56.81 \pm 2.93$   &  $64.99 \pm 3.47$  & $74.56 \pm 0.98$   &  $80.29 \pm 0.90$  \\
& \cellcolor{blue!15}{LP++} & \cellcolor{blue!15}{$\bm{57.23} \pm 1.63$} & \cellcolor{blue!15}{$61.65 \pm 1.66$} & \cellcolor{blue!15}{$68.67 \pm 2.21$} & \cellcolor{blue!15}{$75.86 \pm 0.99$} & \cellcolor{blue!15}{$80.53 \pm 1.05$}\\
\midrule

\multirow{10}{*}{\textit{OxfordPets}} & Zero-shot CLIP$_\text{ ICML'21}$\cite{radford2021clip} & \multicolumn{5}{c}{85.75}\\
\cline{2-7}
& CoOp$_\text{ IJCV'22}$\cite{Zhou2022coop} & $86.27 \pm 1.17$ &  $86.33 \pm 1.13$ & $85.34 \pm 1.69$   & $87.85 \pm 1.21$  & $88.68 \pm 0.71$ \\
& PLOT$_\text{ ICLR'23}$\cite{chen2023plot} & $87.15 \pm 0.72$ & $87.23 \pm 1.21$  &  $88.03 \pm 0.49$  & $88.38 \pm 0.64$  &  $88.23 \pm 0.54$  \\
& KgCoOp$_\text{ CVPR'23}$\cite{Yao2023kgcoop} & $87.51 \pm 0.68$  & $87.51 \pm 0.75$ & $88.04 \pm 0.46$ & $88.59 \pm 0.34$ &  $89.28 \pm 0.21$ \\
& ProGrad$_\text{ ICCV'23}$\cite{Zhu2023prograd} & $\bm{88.34} \pm 0.65$  & $\bm{87.88} \pm 0.69$ & $\bm{88.59} \pm 0.58$ & $\bm{88.87} \pm 0.42$ & $\bm{89.39} \pm 0.47$\\
\cline{2-7}
& CLIP-Adapter$_\text{ IJCV'23}$\cite{Gao2023clipadapter} & $85.46 \pm 0.48 $ & $86.37 \pm 0.25$ & $87.21 \pm 0.51$ & $87.95 \pm 0.26$ & $ 88.33 \pm 0.33$ \\
& Tip-Adapter-F$_\text{ ECCV'22}$\cite{zhang2022tip} & $85.70 \pm 0.16$ & $86.05 \pm 0.46$ & $86.40 \pm 0.29$ & $87.66 \pm 0.28$ & $89.08 \pm 0.27$ \\
& Tip-Adapter-F*$_\text{ ECCV'22}$\cite{zhang2022tip} & $86.05 \pm 0.36$ & $86.49 \pm 0.61$ & $87.19 \pm 0.36$ & $87.89 \pm 0.34$ & $88.26 \pm 0.37$ \\
\cline{2-7}
& Standard LP$_\text{ ICML'21}$\cite{radford2021clip} & $30.62 \pm 1.61$  &  $42.64 \pm 2.03$  &  $55.60 \pm 0.98$  &  $67.32 \pm 0.98$  & $76.23 \pm 0.38$  \\
& \cellcolor{blue!15}{LP++} & \cellcolor{blue!15}{$84.24 \pm 1.36$} & \cellcolor{blue!15}{$85.74 \pm 0.56$} & \cellcolor{blue!15}{$86.94 \pm 0.48$} & \cellcolor{blue!15}{$87.71 \pm 0.65$} & \cellcolor{blue!15}{$88.38 \pm 0.61$}\\
\midrule

\multirow{10}{*}{\textit{Food101}} & Zero-shot CLIP$_\text{ ICML'21}$\cite{radford2021clip} & \multicolumn{5}{c}{77.35}\\
\cline{2-7}
& CoOp$_\text{ IJCV'22}$\cite{Zhou2022coop} & $75.58 \pm 1.29$  &  $77.49 \pm 0.41$ & $77.93 \pm 0.58$  &  $78.92 \pm 0.19$ & $79.21 \pm 0.36$ \\
& PLOT$_\text{ ICLR'23}$\cite{chen2023plot} & $77.46 \pm 0.55$ & $77.72 \pm 0.26$ & $78.23 \pm 0.25$ & $78.40 \pm 0.35$ & $78.86 \pm 0.19$  \\
& KgCoOp$_\text{ CVPR'23}$\cite{Yao2023kgcoop} & $77.20 \pm 0.77$  & $\bm{78.04} \pm 0.18$  & $77.97 \pm 0.28$ & $78.39 \pm 0.40$  &  $78.73 \pm 0.23$ \\
& ProGrad$_\text{ ICCV'23}$\cite{Zhu2023prograd} & $\bm{78.36} \pm 0.41$  & $78.01 \pm 0.70$ & $\bm{78.38} \pm 0.87$ & $\bm{79.11} \pm 0.18$ & $\bm{79.51} \pm 0.23$\\
\cline{2-7}
& CLIP-Adapter$_\text{ IJCV'23}$\cite{Gao2023clipadapter} & $76.93 \pm 0.19 $ & $77.22 \pm 0.15$ & $77.64 \pm 0.17$ & $77.97 \pm 0.22$ & $ 78.45 \pm 0.14$ \\
& Tip-Adapter-F$_\text{ ECCV'22}$\cite{zhang2022tip} & $77.53 \pm 0.14$  & $77.53 \pm 0.22$  & $77.82 \pm 0.27$ & $78.26 \pm 0.22$ & $78.99 \pm 0.15$ \\
& Tip-Adapter-F*$_\text{ ECCV'22}$\cite{zhang2022tip} & $77.58 \pm 0.10$ & $77.36 \pm0.39$ & $77.78 \pm0.15$ & $78.17\pm 0.11$ & $78.72\pm 0.06$ \\
\cline{2-7}
&Standard LP$_\text{ ICML'21}$\cite{radford2021clip} & $31.59 \pm 1.20$  &  $44.60 \pm 1.03$  & $56.13 \pm 0.63$   &  $64.45 \pm 0.55$  &  $70.97 \pm 0.19$  \\
& \cellcolor{blue!15}{LP++} & \cellcolor{blue!15}{$76.61 \pm 0.77$} & \cellcolor{blue!15}{$77.22 \pm 0.55$} & \cellcolor{blue!15}{$77.79 \pm 0.34$} & \cellcolor{blue!15}{$78.53 \pm 0.14$} & \cellcolor{blue!15}{$78.88 \pm 0.19$}\\
\midrule

\end{tabular}
    \vspace{0.1cm}
    \caption*{Table ~\ref{table:resultOther_AllDatasets_details}. \textbf{Comparison to state-of-the-art methods} (Continued). Average classification accuracy (\%) and standard deviation over 10 tasks for 11 benchmarks, 
    Best values are highlighted in bold.}
    
\end{table*}

\end{document}